\documentclass[conference,compsoc]{IEEEtran}
\IEEEoverridecommandlockouts
\usepackage{cite}
\usepackage{amsmath,amssymb,amsfonts}
\usepackage{algorithmic}
\usepackage{graphicx}
\usepackage{textcomp}
\usepackage{xcolor}
\usepackage{hyperref}
\usepackage{clrscode}
\usepackage{multirow}
\usepackage{multicol}
\usepackage{mathtools}
\usepackage{amsfonts}
\usepackage{array}

\usepackage{amsthm}
\usepackage{makecell}
\usepackage{tikz}
\usepackage{pgfplots}
\usepackage{booktabs}
\usepackage{tabularx}
\usepackage{caption}
\usepackage{subcaption}
\usepackage{enumitem}
\usepackage{xspace}

\usepackage{algorithm}
\usepackage{amsthm}
\usepackage{thmtools, thm-restate}
\usepackage{color,soul}

\newcommand{\mbbE}{\mathbb{E}}

\newcommand{\calF}{\mathcal{F}}

\newcommand{\calX}{\mathcal{X}}

\newcommand{\calM}{\mathcal{M}}
\newcommand{\calN}{\mathcal{N}}

\newcommand{\bQ}{\mathbf{Q}}

\newcommand*{\ie}{i.e.\@\xspace}

\declaretheorem{theorem}

\newtheorem{definition}{Definition}
\newtheorem{proposition}{Proposition}

\newcommand{\ourmethod}{\textit{Spectral-DP}\xspace}
\newcommand{\ourmethodblock}{\textit{Block Spectral-DP}\xspace}

\def\BibTeX{{\rm B\kern-.05em{\sc i\kern-.025em b}\kern-.08em
    T\kern-.1667em\lower.7ex\hbox{E}\kern-.125emX}}
\begin{document}




\title{ \bf Spectral-DP: \underline{D}ifferentially \underline{P}rivate Deep Learning through \underline{Spectral} Perturbation and Filtering} 

\author{\IEEEauthorblockN{Ce Feng}
\IEEEauthorblockA{\textit{Lehigh University} \\
cef419@lehigh.edu}
\and
\IEEEauthorblockN{Nuo Xu}
\IEEEauthorblockA{\textit{Lehigh University} \\
nux219@lehigh.edu}
\and
\IEEEauthorblockN{Wujie Wen}
\IEEEauthorblockA{\textit{Lehigh University} \\
wuw219@lehigh.edu}
\and
\IEEEauthorblockN{Parv Venkitasubramaniam}
\IEEEauthorblockA{\textit{Lehigh University} \\
pav309@lehigh.edu}
\and
\IEEEauthorblockN{Caiwen Ding}
\IEEEauthorblockA{\textit{University of Connecticut} \\
caiwen.ding@uconn.edu}
}

\author{
\\[-5.0ex]
\IEEEauthorblockN{Ce Feng$^{*1}$, 
Nuo Xu$^{*1}$, 
\thanks{$^*$These authors contributed equally to this work. }
Wujie Wen$^1$, 
Parv Venkitasubramaniam$^1$, 
Caiwen Ding$^2$
}
\IEEEauthorblockA{\textit{
$^1$Lehigh University,
$^2$University of Connecticut
} \\
\{cef419, nux219, wuw219, pav309\}@lehigh.edu,
caiwen.ding@uconn.edu
} \\
\\[-5.0ex]
}

\maketitle

\begin{abstract}
Differential privacy is a widely accepted measure of privacy in the context of deep learning algorithms, and achieving it relies on a noisy training approach known as differentially private stochastic gradient descent (DP-SGD). DP-SGD requires direct noise addition to every gradient in a dense neural network, the privacy is achieved at a significant utility cost. In this work, we present \ourmethod, a new differentially private learning approach which combines gradient perturbation in the spectral domain with spectral filtering to achieve a desired privacy guarantee with a lower noise scale and thus better utility. 
We develop differentially private deep learning methods based on \ourmethod for architectures that contain both convolution and fully connected layers. In particular, for fully connected layers, we combine a block-circulant based spatial restructuring with \ourmethod to achieve better utility. Through comprehensive experiments, we study and provide guidelines to implement \ourmethod deep learning on benchmark datasets. In comparison with state-of-the-art DP-SGD based approaches, \ourmethod is shown to have uniformly better utility performance in both training from scratch and transfer learning settings.
\end{abstract}


\setcounter{section}{0}

\vspace{-3pt}
\section{Introduction}\label{sec:introduction}
\vspace{-3pt}
Deep Learning algorithms have had tremendous success in a variety of domains in the last several years, due to their ability to extract inferences from data that aid in a variety of tasks. 
Deep learning, however, requires substantial training of several layers densely populated with weight vectors, which is enabled by large datasets often containing sensitive information. As a result, the learned models can be exploited by adversaries to extract the sensitive information in the training datasets. For instance, information about medical procedures can be determined using models built on hospital datasets \cite{shokrietal:2017}. It is therefore critical to provide strong and rigorous privacy guarantees for learning, and in particular, deep learning algorithms.

Over the last several years, different approaches \cite{yu2019differentially,yu2021large,papernot2021tempered, Stevens2022,Andrew2019,Pichapati2019,lee2018concentrated} have been proposed to guarantee \textit{Differential privacy} \cite{DP_algorithm} -- an accepted quantitative measure of privacy-- for learning algorithms, which has been subsequently adapted specifically to deep learning algorithms. These methods invariably rely on a \textit{noisy} training approach known as \textit{Differentially Private Stochastic Gradient (DP-SGD)}, which while privacy preserving, often results in high utility loss. While there have been other advancements to enhance the privacy of deep learning algorithms, they supplement rather than provide an alternative to DP-SGD based noise addition. The main contribution of this work is a new approach to achieving differential privacy in deep learning, called Spectral-DP, which is an alternative to DP-SGD based approaches, and our results will show that this alternative can outperform DP-SGD based approaches.


In the context of deep learning, the fundamental DP-SGD approach, along with its many variants, requires direct noise addition to every weight in a dense neural network, which has a significant impact on utility. Consequently, the more significant improvements to DP-SGD in recent advances have either considered altering specifics of the deep learning architecture, or "curing" datasets, rather than altering the methodology of noise addition. For instance, \cite{papernot2021tempered} explored the use of tempered sigmoid activations to improve the deep learning model's private-learning suitability and achievable privacy-utility tradeoffs (with noise addition through DP-SGD). Yet another approach that improves DP-SGD based methods is to derive handcrafted features (with a data independent preprocessing model) as in \cite{tramer2021differentially}, where it is shown that for a fixed privacy level, deep learning model with handcrafted features outperforms end-to-end deep learning models.

Our approach is motivated by the knowledge that the utility loss in DP-SGD based perturbation methods is consequent to the direct gradient clipping and noise addition at the ``signal" domain of the weights. As a result, the utility is highly sensitive to the noise scale and the clipping norm. Furthermore, there is a tension between using more weights to overcome the effects of noise, and the consequent overfitting that leads to lower utility. Our work here comes out of a hypothesis that although weight vectors have large dimension in the signal or time domain, given the density of the network, they can afford to be sparsely distributed, albeit in a transformed domain, a \textit{spectral domain}. In other words, if the weights are restricted to a subspace in the spectral domain, it is possible to reduce the level of noise required for privacy without necessarily impacting utility. Our approach, referred to as Spectral-DP, is a method that performs a low-bandwidth noise addition in the \textit{Fourier domain} of the weights, and combined with a filtering based dimensionality reduction, we demonstrate that it outperforms DP-SGD in trading utility for privacy. 

Fourier transform is a classical transformation approach, 
and is a unitary and invertible transform, which allows for the learning gradients to be embedded into the spectral domain without impacting the privacy accounting. Furthermore, we note that the frequency components of the weights that have a significant impact on the model outcomes have lower dimensionality than those in the signal domain of the weights. Put another way, forcing weights to fall into a low frequency spectrum provides a way to \textit{compress} the weight representation, and hence serves as a regularizer to prevent loss of utility through overfitting thus overcoming a weakness in existing methods. 
We derive our motivation from empirical studies, such as in \cite{CSIAM-AM-2-484,xu2019training,rahaman2019spectral}, that demonstrate the so-called ``Frequency Principle", wherein deep neural networks tend to fit functions in the low to medium frequencies during training.
In accordance, we develop, test, and demonstrate in this work a Fourier transform based method to provide differentially private low bandwidth noise addition for deep learning architectures.

Specifically, we propose spectral domain based differential privacy for deep learning architectures that can include both convolutional and fully connected layers. Owing to the classical convolution theorem, convolutional layers are more amenable to computation friendly low bandwidth spectral perturbation. The direct adaptation of the spectral DP to fully connected layers, however, is not straightforward. In particular, the high density of weights in fully connected layers and lack of spatially localized features make direct adaptation of the spectral-DP approach challenging. In this regard, we propose an alternative to spectral filtering to reduce the dimensionality of the weights. Specifically, we adapt a spatial compression technique using block circulant matrices \cite{cirCNN_Ding,blockcirculant_1} which we combine with our Fourier based noise addition approach to develop a compressed spectral domain differentially private training methodology, Block-Spectral DP. As our results will show, in networks with fully connected layers, Block Spectral DP outperforms DP-SGD based approaches.

The overarching contribution of our work is a viable alternative to DP-SGD for deep learning with differential privacy. In particular, we propose approaches that combine spectral noise addition with dimensionality reduction to achieve better utility for a given differential privacy guarantee. Our specific contributions are as follows:
\begin{itemize}[leftmargin=*]
  \setlength\itemsep{0em}
   \item We address a critical challenge in achieving differential privacy in deep learning algorithms which is to reduce the noise scale to achieve better utility. 
    \item Through theoretical analysis, we develop the spectral filtering based noise scale reduction technique, and provide the analytical reasoning for the improved utility performance of our methodologies.
    \item We develop differentially private deep learning algorithms based on our \ourmethod approach for a general class of neural network architectures. Specifically, for convolutional layers our approach combines filtering and spectral gradient perturbation to achieve the desired noise scale reduction.
    \item For fully connected layers, we develop a variant of our fundamental approach, block \ourmethod, where we adapt a spatial compression mechanism using block circulant matrices to the spectral gradient perturbation which further reduces the impact of differentially private noise addition on the utility. 
    \item Through comprehensive experimental study, we provide guidelines to choose the right parameters including filtering ratio and clipping norms to achieve the best privacy utility tradeoff using Spectral-DP. 
    \item Through several experiments on three benchmark image classification datasets, namely MNIST, CIFAR10 and CIFAR100, we demonstrate that \ourmethod can outperform the state-of-the-art implementation of DP-SGD. Specifically, \ourmethod incurs less than 1\% accuracy drop for privacy budget as low as $(2,10^{-5})$ for MNIST and 20\% higher accuracy than DP-SGD for CIFAR10 with privacy budget $(3,10^{-5})$ for training from scratch models. 
    In the transfer learning setting, \ourmethod achieves 94.85\% for CIFAR10 and 77.52\% for CIFAR100 with privacy budget $(1,10^{-5})$.
     Moreover, when combined with Scatter-net based data curation, \ourmethod incurs less than 1\% accuracy loss with privacy budget $(3,10^{-5})$.
\end{itemize}
The remainder of the paper is organized as follows. We provide some preliminaries in Section \ref{sec:prelim}. We formulate the mathematical model and related formulations of \ourmethod in Section \ref{sec:approach}. We conduct several experiments and analyze \ourmethod in Section \ref{sec:experiment}, and discuss the limitations of \ourmethod in Section \ref{sec:limitations}. In Section \ref{sec:relatedwork}, we detail the related work to place our work in broader scientific context. Some concluding remarks are presented in Section \ref{sec:conclusion}.

\vspace{-4pt}
\section{Preliminary}\label{sec:prelim}
\vspace{-4pt}
In this section, we first provide a brief background on differential privacy. We then introduce the basics of differentially private stochastic gradient descent (DG-SGD) for privacy-preserving deep learning model training, followed by its R{\'{e}}nyi differential privacy (RDP) based version for the tightest privacy account analysis on DP-SGD.


\begin{figure*}[t]
    \centering
    \includegraphics[width=1.0\linewidth]
    {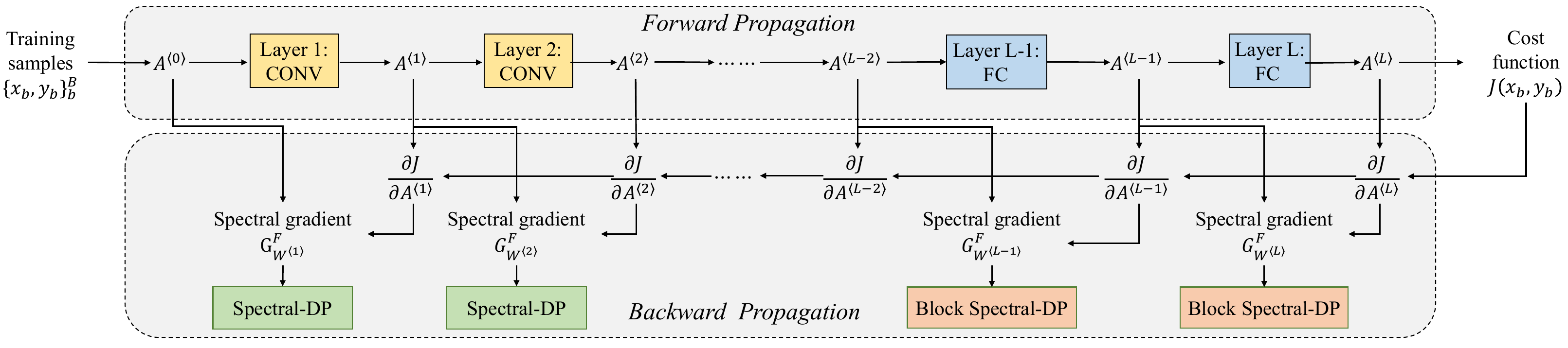}
    \caption{The backpropagation process of \ourmethod based private training for deep learning models.}
    \label{fig:spectralDP_overview}
\vspace{-15pt}
\end{figure*}

\vspace{-4pt}
\subsection{Differential Privacy}
\vspace{-4pt}
Differential privacy \cite{DP_algorithm} is a quantitative definition of privacy, initially designed in the context of databases. Specifically, for two adjacent databases - \ie databases that differ only on a single entry - differential privacy achieved by an algorithm $\calM$ is formally defined as follows:
Differential privacy \cite{DP_algorithm} is a privacy definition that describes the privacy loss associated with application that utilizes from a database. It is defined on adjacent databases. We say two databases are adjacent if they differ only in a single entry. Hence, the different privacy is defined by
\begin{definition}\label{def:dp}
    A randomized algorithm $\mathcal{M}$ with domain $\mathcal{D}$ and $\mathcal{R}=\text{Range}(\mathcal{M})$ is $(\epsilon, \delta)-$differentially private if for all $S\subseteq\mathcal{R}$ and for any two adjacent sets $d,d'\in\mathcal{D}$:
    \begin{equation}\label{eq:dp_definition}
        \text{Pr}[\mathcal{M}(d)\in S]\leqslant \exp ^\epsilon \text{Pr}[\mathcal{M}(d')\in S] + \delta
    \end{equation}
\end{definition}
A prevalent technique for designing a differentially private mechanism is to add controlled noise from specific distribution. The noise level is controlled by the sensitivity of a function $f:\mathcal{D}\mapsto\mathbb{R}$. For instance, Gaussian mechanism takes as input the function $f$ with sensitivity $S$ and a parameter $\sigma$ controlling the Gaussian noise.  The Gaussian mechanism is formulated as noise perturbation in the output of the sequence:
\begin{equation}\label{eq:GaussianMechanism}
\setlength\abovedisplayskip{3pt}
\setlength\belowdisplayskip{3pt}
    Gauss(f,S,\sigma) \triangleq f + \mathcal{N}(0,S^2\cdot\sigma^2)
\end{equation}
where $\mathcal{N}(0,S^2\cdot\sigma^2)$ is the Gaussian distribution with zero mean and variance $S^2\cdot\sigma^2$. It is noticed that the noise scale in Gaussian mechanism is proportional to the sensitivity. And the sensitivity of $f$ is the maximum of the absolute distance $|f(d)-f(d')|$ where $d$ and $d'$ are adjacent inputs. According to Theorem 3.22 in \cite{DP_algorithm}, for any $\epsilon,\delta\in(0,1)$, the Gaussian mechanism achieve $(\epsilon,\delta)-$differential privacy when $\sigma=\sqrt{2\log(1.25/\delta)}/\epsilon$.

\vspace{-3pt}
\subsection{Differential Privacy in Deep Learning}\label{subsec:Background_DPSGD}
\vspace{-3pt}
In the context of deep learning, we require that the algorithm that produces the learned model is $(\epsilon,\delta)-$ differentially private with respect to the training dataset. In that regard, the most common approaches utilize the idea of differentially private stochastic gradient descent (DP-SGD) \cite{dpsgd}.
DP-SGD introduces differential privacy into deep learning by controlling the influence of the training data during the training process. Specifically in each training iteration, the per-example gradients $g_i$ are first computed with respect to the cost function $J$. The gradients are clipped according to some predefined threshold $C$. The key to achieving privacy is to add Gaussian noise to the average of the clipped per-example gradients directly, wherein the noise level $\sigma^2C^2$ is proportional to the $L_2$ sensitivity of the average gradient. 

Since training a deep learning model occurs over several iterations, Gaussian noise is added at every iteration, and the overall differential privacy is computed through a composition or a privacy accountant mechanism. The best known privacy accounting is based on a modification of the differential privacy definition, known as R{\'{e}}nyi Differential Privacy (RDP). RDP is defined in Definition \ref{def:RDP}.

\begin{definition}\label{def:RDP}
    A randomized algorithm $\mathcal{M}$ with domain $\mathcal{D}$ and $\mathcal{R}=\text{Range}(\mathcal{M})$ is $(\epsilon, \delta)-$RDP if for any two adjacent sets $d,d'\in\mathcal{D}$:
    \begin{equation}\label{eq:RDP_definition}
    \setlength\abovedisplayskip{3pt}
    \setlength\belowdisplayskip{3pt}
        D_{\alpha}(\calM(d)\|\calM(d'))\leqslant \epsilon
    \end{equation}
\end{definition}
where $D_{\alpha}(\cdot\|\cdot)$ is the R{\'{e}}nyi divergence between two probability distributions. More detailed, it is defined by
\begin{definition}\label{def:RenyiDiver}
    For two probability distributions $P$ and $Q$ defined on $\calX$ over the same space, and let $p$ and $q$ denote the densities of $P$ and $Q$, respectively, the  R{\'{e}}nyi divergence of order $\alpha>1$ is given by
    \begin{equation}\label{eq:Renyi-divergence_definition}
    \setlength\abovedisplayskip{3pt}
    \setlength\belowdisplayskip{3pt}
        D_{\alpha}(P\|Q)\triangleq\frac{1}{\alpha-1}\log\mathop\mbbE\limits_{x\thicksim Q}(\frac{P(x)}{Q(x)})^{\alpha}
    \end{equation}
\end{definition}
To analyze the composition of DP-SGD, the DP guarantee of a single iteration of DP-SGD is firstly converted into its equivalent RDP using Proposition \ref{prop:RDP_conversion} in \cite{RDP2017}, and subsequently for $T$ training iterations, the total RDP guarantee can be obtained by Proposition \ref{prop:RDP_composition} in \cite{RDP2017}. Finally, the total DP guarantee can be converted back from the total RDP guarantee by Proposition \ref{prop:RDP_conversion}. 
\begin{proposition}\label{prop:RDP_conversion}
    (From RDP to $(\epsilon,\delta)$-DP). If $f$ is an $(\alpha,\epsilon)-RDP$ mechanisms, it also satisfies $(\epsilon+\frac{\log 1/\delta}{\alpha-1},\delta)$-differential privacy for any $0<\delta<1$.
\end{proposition}
\begin{proposition}\label{prop:RDP_composition}
    For a $(\alpha,\epsilon_1)-RDP$ mechanism $f$ and a  $(\alpha,\epsilon_2)-RDP$ mechanism $g$, then the mechanism $f\circ g$ satisfies $(\alpha,\epsilon_1+\epsilon_2)-RDP$.
\end{proposition}
Although our approach is different from DP-SGD wherein noise is not added directly to the gradients, we will utilize the above propositions by drawing an equivalence to differential privacy in each iteration, and subsequently using the RDP based privacy accounting to compute the total differential privacy of our approach.

\vspace{-3pt}
\section{Approach}\label{sec:approach}
\vspace{-3pt}
In this section, we formally present \ourmethod in the context of deep learning. While \ourmethod shares the same objective with DP-SGD which aims to perturb weight gradients during the training process, it can principally reduce the differential privacy (DP) noise, thus to significantly escalate model utility, by conducting dedicated spectral filtering in our designed DP noise addition process performed in the spectral domain, with theoretical guarantee on the privacy.

Figure \ref{fig:spectralDP_overview} provides an overview of our approach. Consider a deep learning model consisting of $L$ layers (either convolutional or fully connected). During the backward propagation process of the learning algorithm, the gradients at each layer are transformed into their spectral representation which is subsequently perturbed by Gaussian noise, and filtered, prior to transforming back to the signal (or spatial) domain. Since a convolutional operation in the spatial domain is equivalent to multiplication in the spectral domain, \textbf{convolutional (CONV) layers} are more amenable to spectral gradient perturbation. For \textbf{fully connected (FC) layers}, we supplement this mechanism with a block-circulant matrix based weight compression and restructuring to address the high density of weights prior to spectral perturbation and filtering.

The remainder of the section is organized as follows. In Section \ref{sec:FourierDP}, we present the conceptual basis of \ourmethod and a theoretical analysis that provides the differential privacy guarantee of the method, and demonstrates the reduction in noise scale that enables the better utility performance of spectral perturbation and filtering. In Sections \ref{sec:SpectralFourierDP} and \ref{sec:blockFourierDP}  we describe in detail the spectral perturbation and filtering methodology as applied to convolutional layers and fully connected layers respectively. In Section \ref{sec:DPtogether}, we outline the overall training of a neural network with both kinds of layers over multiple iterations to achieve a desired guarantee of $(\epsilon, \delta)$ differential privacy. 

\vspace{-3pt}
\subsection{Conceptual Foundations of Spectral-DP}\label{sec:FourierDP}
\vspace{-3pt}
In this section, we present the concept and theoretical analysis of \ourmethod which is the basis of the specific deep learning algorithms developed in subsequent sections.

\subsubsection{\ourmethod Overview}
The key of \ourmethod is to perturb weight gradients in the spectral or Fourier domain by taking advantage of existing primitives such as Fourier transform and the Gaussian mechanism for differential privacy. Specifically, Fourier transform (FT) is used to project data to the spectral domain (or frequency domain), and the algorithm perturbs the Fourier transform coefficients prior to filtering out a fraction of the coefficients. The approach is described in mathematical detail below.

Consider an $N$ length sequence $\bQ = \{Q_0,Q_1,\cdots,Q_{N-1}\}$ as an example. We denote $\{F^N\}:=\{F_0, F_1,\cdots,F_{N-1}\}$ as a collection of all spectral coefficients of $\bQ$, where each $F_i$ is computed by:
\setlength\abovedisplayskip{3pt}
\setlength\belowdisplayskip{3pt}
$$F_i=\frac{1}{\sqrt{N}}\sum_{n=0}^{N-1}Q_n\cdot e^{-\frac{j2\pi}{N}in}$$

The Gaussian noise addition mechanism is applied into $\{F^N\}$. Since the Gaussian noise scale is proportional to the $L_2$ norm of the $\{F^N\}$, we bound $\{F^N\}$ by a clipping parameter $S$: 
\setlength\abovedisplayskip{3pt}
\setlength\belowdisplayskip{3pt}
$$\bar{F}^N=F^N/ \max\{1,\frac{\|F^N\|_2}{S}\}$$
where $\|\cdot\|_2$ denotes the $L_2$ norm. The spectral coefficients are perturbed with additive Gaussian noise:
$$\tilde{F}^N=\bar{F}^N+V^N$$
where $V^N=\{V_0,V_1,\cdots,V_{N-1}\}$ is the noise vector, and each $V_i$ is drawn from $\calN(0,\sigma^2S^2)$ independently. We denote this process as $Gauss(\bar{F}^N,S,\sigma)$. A key mechanism that allows \ourmethod to limit the impact of noise is filtering, in other words, eliminating a fraction of the coefficients:
$$P_K(\tilde{F}_i)=\left\{\begin{array}{lc}
        \tilde{F}_i & \text{if }i<K \\
        0 & \text{otherwise}
    \end{array}\right.
$$
the $K/N$ determines the fraction of coefficients which are perturbed and allows us to reduce the overall noise scale. Our motivation is that it is sufficient to concentrate the weights in a \textit{low bandwidth} space without compromising on utility while saving on the impact of noise. The perturbed and filtered coefficients are retransformed to the signal domain using the inverse Fourier Transform (I-FT). 
The overall procedure is outlined in Algorithm \ref{alg:FourierDP}.

\begin{algorithm}[t]
    \caption{\ourmethod perturbation}
    \label{alg:FourierDP}
  \begin{algorithmic}[1]
    \REQUIRE Query $Q=\{Q_0,Q_1,\cdots,Q_{N-1}\}$, $l_2$ sensitivity $S$, noise scale $\sigma$\\
    \textbf{Output: } $\tilde{Q}$\\
    \STATE Compute the Fourier coefficients of $Q$
    \STATE Clipping the Fourier coefficients by $S$
    \STATE Noise addition:
        $\tilde{F}^N=Gaussian(\bar{F^N}, S,\sigma)$
    \STATE Spectral filtering:
    $\hat{F}_i^K=P_K(\tilde{F}_i)
    $
    \STATE Inverse Fourier transformation:
    $\tilde{Q_n}=\text{I-FT}(\hat{F_i}^K)$
  \end{algorithmic}
\end{algorithm}
\setlength{\textfloatsep}{10pt}

\subsubsection{Theoretical analysis of \ourmethod}
\label{subsec:theoreticalanalysis}
In the following theorem, we determine the privacy budget of Algorithm \ref{alg:FourierDP}, and prove that spectral perturbation achieves the desired differential privacy.

\begin{restatable}{theorem}{Privacythm}
\label{thm:FourierDP}
    In Algorithm \ref{alg:FourierDP}, the output $\tilde{Q_n}$ is ($\epsilon,\delta$) differentially private if we choose $\sigma$ to be $\sqrt{2\log(1.25/\delta)}/\epsilon$.
\end{restatable}

\begin{proof}
    The proof relies on Theorem 3.22 in \cite{DP_algorithm} and the post-processing property of DP algorithm. The detailed proof is given in Appendix~\ref{Appendix:proof_privacythm}.
\end{proof}

Since both spectral filtering and inverse Fourier transformation can be treated as the post-processing steps that do not alter the DP budget, \ourmethod better utilizes the privacy budget. As demonstrated in the following Proposition, the filtering operation in spectral domain leads to prominent noise scale reduction.
\begin{restatable}{proposition}{PropNoiseReduction}
\label{prop:noise_reduction}
    Let $V^N=\{V_0,V_1,\cdots,V_i,\cdots,V_{N-1}\}$ be a collection of noise vector in spectral domain, and each $V_i$ is drawn from $\calN(0,\sigma^2S^2)$. Consider $v_n=\text{I-FT}(P^K(V_i))$, then $v_n$ follows a normal distribution $\calN(0,\frac{K}{N}\sigma^2S^2)$.
\end{restatable}
\begin{proof}
\setlength\abovedisplayskip{3pt}
\setlength\belowdisplayskip{3pt}
    The detailed proof is given in Appendix \ref{Appendix:noise_reduction}.
\end{proof}

As Proposition \ref{prop:noise_reduction} shows, the spectral filtering allows the reduction in the overall noise scale from $\sigma^2S^2$ down to $\frac{K}{N}\cdot\sigma^2S^2$, where $K<N$. 
Consequently, should the filtering mechanism not affect the utility, the noise reduction could significantly minimize the utility penalty incurred by differential private training. While filtering more frequency components (a smaller $K$) indicates more DP noise reduction thus less utility penalty by DP, the more weight distortion errors after Inverse Fourier Transform could inevitably impact the gained model utility. We define a key parameter--\textbf{filtering ratio} $\rho=(K-N)/N$, to balance the impact of these two factors and 
will discuss the impact and choice of this key parameter in Sections~\ref{subsec:effectiveness_spectral_dp} and \ref{subsec:block_spectral_dp}

\begin{figure*}[h]\label{fig:method_overview}
     \centering
     
     \begin{subfigure}[b]{0.95\linewidth}
         \centering
         \includegraphics[width=\textwidth]{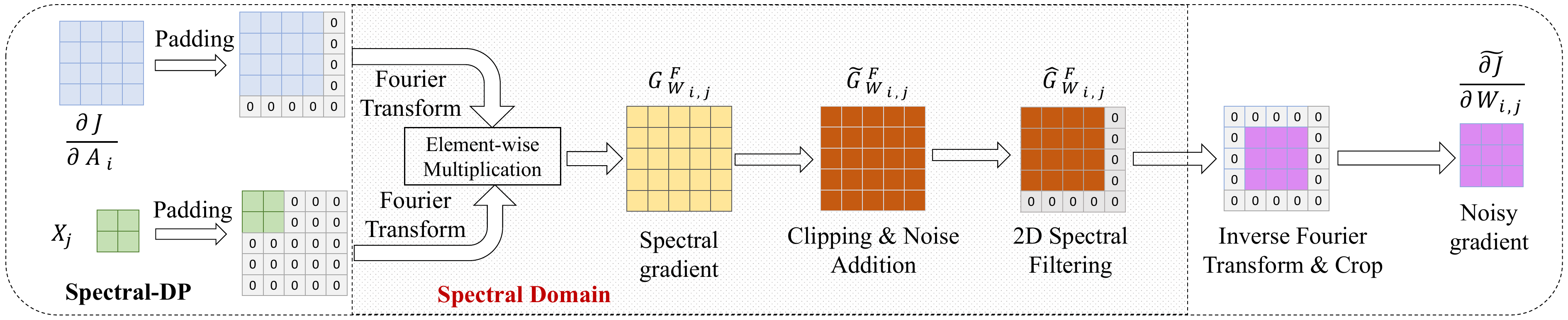}
         \caption{\ourmethod implemented into a convolutional layer.}
         \label{subfig:spectral_detail}
     \end{subfigure}
     
     
     \begin{subfigure}[b]{0.95\linewidth}
         \centering
         \includegraphics[width=\textwidth]{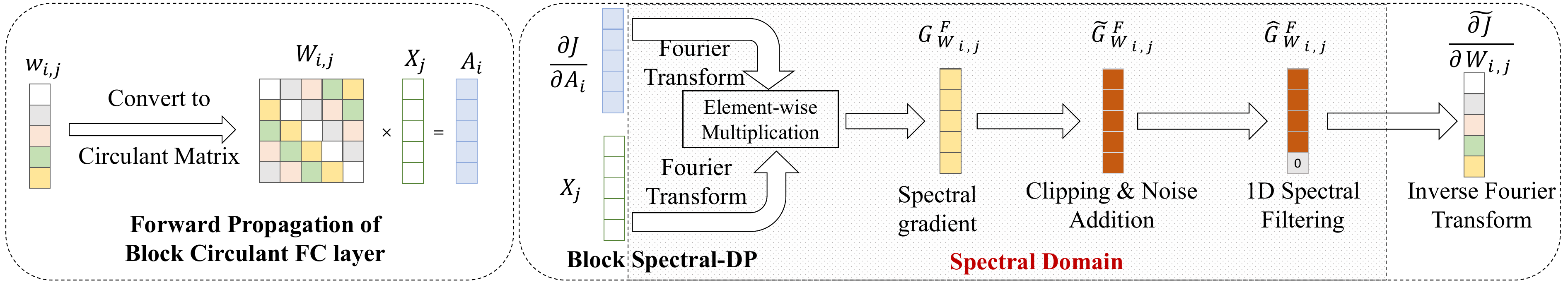}
         \caption{\ourmethodblock implemented into a block circulant fully connected layer. }
         \label{subfig:block_spectral_detail}
     \end{subfigure}
     \caption{Detailed Spectral-DP framework. 
     }
     \label{fig:method_overview}
\vspace{-10pt}
\end{figure*}
\subsection{\ourmethod in CONV Layer }\label{sec:SpectralFourierDP}
\subsubsection{Adapting \ourmethod to 2D CONV}
\label{subsec:adaptingspectralDPin2dconv}
To adapt \ourmethod for private deep learning, our first question would be how to perform gradient perturbations for different types of model layers using \ourmethod. We first focus on the 2-dimensional (2D) convolution that dominates the operations of the convolutional layer.

According to the convolution theorem, the 2D convolution in spatial domain can be easily converted into element-wise multiplication of two matrices in the spectral domain. 
\ourmethod is then applied into the element-wise multiplication and mainly consists of a Gaussian noise addition and a 2D spectral filtering. 
To demonstrate how effectively \ourmethod reduces DP noises in 2D convolution, we further derive the relation between the noise scale and filtering parameter $\rho$ in Corollary \ref{prop:noise_reduction_2D}. 
\begin{restatable}{corollary}{PropNoiseReductionConv}
\label{prop:noise_reduction_2D}
    Let $V^N$ be the collection of a noise vector $\{V_{i,j}\}$ where $i\in\{0,1,\cdots,N-1\}$ and $j\in\{0,1,\cdots,N-1\}$ in spectral domain, and
    each $V_{i,j}$ be drawn from $\calN(0,\sigma^2S^2)$, consider a 2D spectral filtering:
    $$P_{2D}^K=\left\{
    \begin{array}{cl}
    V_{ij}     & \text{if }i<K\text{ and } j<K \\
    0     & \text{otherwise}
    \end{array}\right.$$
    and $v_{mn}=\mathcal{F}^{-1}(P_{2D}^K(V_{i,j}))$, then $v_{mn}$ follows a normal distribution $\calN(0,\frac{K^2}{N^2}\sigma^2S^2)$.
\end{restatable}
\begin{proof}
    The detailed proof is given in Appendix \ref{Appendix:noise_reduction2D}.
\end{proof}

\subsubsection{Adapting \ourmethod into CONV layer}
\label{subsec:conv_spectral_dp}
\begin{algorithm}[t]
    \caption{\ourmethod in a 2D convolutional layer}
    \label{alg:SpectralFourierDP}
  \begin{algorithmic}[1]
    \REQUIRE A CONV layer with input $X$, filters $W$, $\frac{\partial{J}}{\partial{A}}$, clipping bound $S$, $\sigma$, filtering ratio $\rho\in(0,1)$ \\
    \textbf{Output:} $\tilde{\frac{\partial J}{\partial W}}$\\
    $\bullet$  \textbf{\textit{Stage I:}} Noise addition
    \STATE \textbf{Compute:} \\$G_W^F=\{G_{W_{i,j}}^F\}$, where $i\in\{1,\cdots,C_{out}\}$, $j\in\{1,\cdots,C_{in}\}$.\\
    \STATE Clipping Norm: $\bar{G}_W^F=G_W^F/\max(1,\frac{\|G_w^F\|_2}{S})$
    \STATE \textbf{Gaussian mechanism:} \\$\tilde{G}_W^F= Gauss(\bar{G}_W^F,S,\sigma)$\\
    $\bullet$  \textbf{\textit{Stage II:}} Filter-wise pruning and inverse Fourier Transform\\
    \FOR {$i\in 1,2,\cdots,C_{out}$}
    \FOR {$j\in 1,2,\cdots,C_{in}$}
    \STATE \textbf{2D Spectral filtering:} \\$\hat{G}_{W_{i,j}}^F\leftarrow$Zero last $\rho$ of rows and columns in $\tilde{G}_{W_{i,j}}^F$ 
    \STATE \textbf{Inverse Fourier transform:}\\ $\tilde{\frac{\partial J}{\partial W_{i,j}}}=\mathcal{F}^{-1}[[\hat{G}_{W_{i,j}}^F]_0]$
    \ENDFOR
    \ENDFOR
    \STATE $\tilde{\frac{\partial J}{\partial W}}=\{\tilde{\frac{\partial J}{\partial W_{i,j}}}\}$, $i\in\{1,\cdots,C_{out}\}$, $j\in\{1,\cdots,C_{in}\}$.
  \end{algorithmic}
\end{algorithm}
\setlength{\textfloatsep}{5pt}

Based on the 2D spectral filtering, we then provide more implementation details of \ourmethod for training convolutional layers. We consider a typical 2-dimensional convolutional layer of a deep learning model with the input vector $X\in\mathbb{R}^{H_{in}\times W_{in}\times C_{in}}$ and convolution filters $W\in\mathbb{R}^{C_{out}\times C_{in}\times d\times d}$, where $C_{in}$ and $C_{out}$ are the number of input and output channels, $d\times d$ is the size of a 2D convolution filter. The forward propagation process of the inference in the layer is expressed as follow (bias and activation are omitted):
$$A_{i}=\sum_{j}^{C_{in}}X_{j}\circledast W_{i,j}$$
where $A_i$ denotes the $i-$th channel of the convolution output 
with size $H_{out}\times W_{out}$ where $H_{out}=H_{in}+d-1$ and $W_{out}=W_{in}+d-1$ 
, $X_j$ denotes the $j-$th channel of the input, $W_{i,j}$ denotes the convolution filter that connects the $i-$th channel of the output and the $j-th$ channel of the input, and $\circledast$ denotes the 2D convolution.
For each convolution filter, the backward propagation and the convolution theorem indicate that the gradient can be approximately expressed as
\begin{equation}\label{eq:conv_grad}
    \frac{\partial J}{W_{i,j}}=\frac{\partial J}{\partial A_i}\circledast\frac{\partial A_i}{\partial W_{i,j}}=\mathcal{F}^{-1}[\mathcal{F}[\frac{\partial J}{\partial A_i}] \odot \mathcal{F}[X_j]]
\end{equation}
where $\frac{\partial J}{W_{i,j}}$ denotes the gradient of $W_{i,j}$ w.r.t. the cost function $J$, $\frac{\partial J}{\partial A_i}$ denotes the gradient of $A_i$ w.r.t. $J$, $\frac{\partial A_i}{\partial W_{i,j}}$ denotes the gradient of $W_{i,j}$ w.r.t. $A_i$, $\mathcal{F}$ and $\mathcal{F}^{-1}$ are the Fourier Transform and inverse Fourier Transform operator, respectively, and $\odot$ denotes the element-wise multiplication. Let $G_{W_{i,j}}^F=\mathcal{F}[\frac{\partial J}{A_i}] \odot \mathcal{F}[X_j]$ be the spectral gradient of $\frac{\partial L}{W_{i,j}}$, then \ourmethod can be directly applied into the spectral gradient. 
Computing the spectral gradient requires a complexity of $O((H_{out}*W_{out})\log(H_{out}*W_{out}))$ where conventional convolution of computing $\frac{\partial J}{W_{i,j}}$ requires a complexity of $O(H_{out}*W_{out}*H_{in}*H_{in})$. Theoretically, the spectral gradient computation is faster than the conventional convolution if $\log(H_{out}*W_{out})<(H_{in}*H_{in})$. We provide detailed complexity analysis in the Appendix \ref{apdx:complexity}.

Figure \ref{fig:method_overview}(a) depicts the implementation steps when applying \ourmethod to the gradient perturbation of a single 2D convolution filter. $\frac{\partial J}{A_i}$ and $X_j$ are first padded to the same size and transformed to the spectral gradient $G^F_{W_{i,j}}$. Consequently, the noisy spectral gradient $\tilde{G}^F_{W_{i,j}}$ is obtained by applying clipping and Gaussian noise addition into $G^F_{W_{i,j}}$.
The filtered spectral gradient $\hat{G}^F_{W_{i,j}}$ is then computed with a filtering ratio $\rho$ using the 2D spectral filtering approach mentioned in Corollary \ref{prop:noise_reduction_2D}.

The main procedure of \ourmethod in 2D convolutional layer is outlined in Algorithm \ref{alg:SpectralFourierDP}.
As the 2D convolutional layer usually contains multiple filters, we denote $G_W^F=\{G_{W_{i,j}}^F\}$ as the spectral gradients of $W$ with respect to the cost function $J$. As shown at the stage I of Algorithm \ref{alg:SpectralFourierDP}, by applying Gaussian mechanism into $G_W^F$, the differential privacy of all parameters in the layer is guaranteed.
At stage II, the spectral domain filtering is applied within each convolution filter. According to Corollary \ref{prop:noise_reduction_2D}, \textit{the 2D spectral filtering provides larger noise reduction than the 1D spectral filtering but leads to larger weight distortion errors.} In Section \ref{sec:experiment}, we conduct comprehensive experiments to evaluate how the filtering ratio affects the utility. 

\vspace{-5pt}
\subsection{Block Spectral-DP in FC Layer}\label{sec:blockFourierDP}
\vspace{-5pt}

In addition to fitting \ourmethod into CONV layers, our next question would be how to extend it to the fully connected (FC) layers. 
The weight matrix in an FC layer often has a much higher dimension than CONV layers which have a weight-sharing mechanism. Furthermore, unlike CONV layers, operations in FC layers cannot directly map to multiplication in the spectral domain. 
To address these, our key idea is to compress and restructure the weight matrix to facilitate the adoption of \ourmethod to FC layers. In this regard, the structure of a block circulant weight matrix~\cite{cirCNN_Ding,blockcirculant_1} is a suitable choice.  Each row vector in such a matrix is the circular shift form of the previous row, and the matrix vector multiplication in time domain can be simplified as vector-vector multiplication in the spectral domain. We further name this approach as \ourmethodblock.   
As we shall show later, \ourmethodblock not only mathematically supports the spectral transformation for adding gradient perturbation, but also compresses redundant weights in FC layers without impacting the utility.
\vspace{-8pt}
\subsubsection{Block circulant based FC layer}
We first introduce the definition of a block circulant matrix. Given a matrix $W$ of size $m\times n$, it is said that $W$ is block circulant if $W$ can be partitioned into $p\times q$ square blocks of circulant matrix $W_{ij}\in\mathbb{R}^{d\times d}$, where $d$ is defined as the block size (size of each sub-matrix block), $p=m\div d$, $q=n\div d$, $i\in\{1,2,\cdots,p\}$, and $j\in\{1,2,\cdots,q\}$. And each square block matrix $W_{i,j}$ is circulant as specified below:
$$\left [ \begin{array}{cccc}
    w_0 & w_1 & \cdots & w_{k-1} \\
    w_{k-1} & w_0 & \cdots & w_{k-2} \\
    \ddots & \ddots & \ddots & \ddots\\
    w_1 & w_2 & \cdots & w_0 \\
\end{array} \right ]$$
We note the circulant matrix can be represented by a vector $w=\{w_0, w_1, \cdots, w_{k-1}\}$
Consider a fully connected layer consisting of $m$ outputs and $n$ inputs and a block circulant weight matrix $W$.
Assume $W$ is partitioned into $p\times q$ blocks of circulant matrix, the forward propagation in the block circulant weight matrix based fully connected layer is given by:
\begin{equation}\label{eq:forward_block}
    A = WX = \left[
    \begin{array}{c}
         \sum_{j=1}^q W_{1,j}X_j \\
         \sum_{j=1}^q W_{2,j}X_j \\
         \vdots\\
         \sum_{j=1}^q W_{p,j}X_j 
    \end{array}
    \right]=\left[
    \begin{array}{c}
         A_1  \\
         A_2 \\
         \vdots \\
         A_p
    \end{array}
    \right]
\end{equation}
where the input $X$ is partitioned as $X=[X_1^T,X_2^T,\cdots,X_q^T]^T$, $A_i\in\mathbb{R}^k$ is a column vector that is the respective output of $\sum_{j=1}^q W_{i,j}x_j$. We further assume each square block sub-matrix $W_{i,j}$ is represented by a vector $w_{i,j}$ where $w_{i,j}$ is the first row of $W_{i,j}$, then according to the \textit{circulant convolution theorem} \cite{Pan2001Structured, Eberly1996Polynomial}, the computation of $W_{i,j}X_j$ can be expressed as $A_i=w_{i,j}*X_j=\calF^{-1} (\calF(w_{i,j})\odot \calF(X_j))$, where $*$ is the operator of circulant convolution. This process is shown in the left block of Figure \ref{fig:method_overview}(b).

We then consider the backward propagation training of the fully connected layer with the block circulant weight matrix. Let $J$ denote the cost function, and $A_{il}$ be the $l-th$ element in $A_i$, by the chain rule, the backward propagation process is derived as
\begin{equation}\label{eq:backward_block}
\setlength\abovedisplayskip{3pt}
\setlength\belowdisplayskip{3pt}
    \frac{\partial J}{\partial w_{i,j}} = \sum_{l=1}^k \frac{\partial J}{A_{i,l}} \frac{\partial A_{i,l}}{w_{i,j}} = \frac{\partial J}{A_{i}} \frac{\partial A_i}{\partial w_{i,j}}
\end{equation}
We note $A_i$ is the circular convolution of $w_{ij}$ and $x_j$ which indicates that $\frac{\partial A_i}{\partial w_{ij}}$ is block circulant matrix. Hence, the computation of Eq.~(\ref{eq:backward_block}) can be expressed as the "Fourier Transform $\longrightarrow$ Element-wise Multiplication $\longrightarrow$ Inverse Fourier Transform". 
The complexity analysis of computing the gradient is provided in Appendix \ref{apdx:complexity}. 

\vspace{-5pt}
\subsubsection{Implementing \ourmethodblock into FC layer}
\begin{algorithm}[t]
    \caption{\ourmethodblock in a single FC layer}
    \label{alg:blockFourierDP}
  \begin{algorithmic}[1]
    \REQUIRE A block circulant weight matrix based FC layer with input $X$ and block circulant matrix $W$. $\frac{\partial{J}}{\partial{A}}$, $\{w_{i,j}\}$, $p$, $q$, block size $d$, $m$, $n$, clipping bound $S$,$\sigma$ filtering parameter $\rho\in(0,1)$ \\
    \textbf{Output: } 
    noisy gradient $\tilde{\frac{\partial{J}}{\partial{W}}}$\\
    $\bullet$  \textbf{\textit{Stage I:}} Noise addition
    \STATE \textbf{Compute:} \\$G_W^F=\{G_{W_{i,j}}^F\}$, where $i\in\{1,2,\cdots,p\}$, $j\in\{1,2,\cdots,q\} $.\\
    \STATE Clipping Norm: $\bar{G}_W^F=G_W^F/\max(1,\frac{\|G_w^F\|_2}{S})$
    \STATE \textbf{Gaussian mechanism:} \\$\tilde{G}_W^F= Gauss(\bar{G}_W^F,S,\sigma)$\\
    $\bullet$  \textbf{\textit{Stage II:}} Block-wise pruning and inverse Fourier Transform\\
    \FOR {$i\in 1,2,\cdots,p$}
    \FOR {$j\in 1,2,\cdots,q$}
    \STATE \textbf{1D Spectral filtering:} \\$\hat{G}_{W_{i,j}}^F\leftarrow$Zero last $\rho$ of coefficients in $\tilde{G}_{W_{i,j}}^F$ 
    \STATE \textbf{Inverse Fourier transform:}\\ $\tilde{\frac{\partial J}{\partial W_{i,j}}}=\mathcal{F}^{-1}[\hat{G}_{W_{i,j}}^F]$
    \ENDFOR
    \ENDFOR
    \STATE $\tilde{\frac{\partial J}{\partial W}}=\{\tilde{\frac{\partial J}{\partial W_{i,j}}}\}$, $i\in\{1,\cdots,p\}$, $j\in\{1,\cdots,q\}$.
  \end{algorithmic}
\end{algorithm}
\setlength{\textfloatsep}{6pt}

We now demonstrate how to implement \ourmethodblock into the fully connected layer. Eq.~(\ref{eq:backward_block}), 
allows us to apply \ourmethodblock into the spectral gradients of the parameters. Then for each square block, the spectral gradient is computed by
\begin{equation}
\setlength\abovedisplayskip{3pt}
\setlength\belowdisplayskip{3pt}
    G^F_{w_{i,j}} = \calF(\frac{\partial J}{\partial A_i})\circ \calF(\frac{\partial A_i}{\partial w_{i,j}})
\end{equation}

Algorithm \ref{alg:blockFourierDP} outlines \ourmethodblock as applied to a fully connected layer. We denote $G_W^F$ as the spectral gradient of $W$ with respect to the cost function $J$. At stage I, the Gaussian mechanism is applied into the $G_W^F$ to ensure the differential privacy guarantee of the spectral gradients. Operations at stage II such as spectral filtering operation and inverse Fourier Transform are introduced as the post-processing of the Gaussian mechanism. We note operations are applied at the sub-matrix level. Unlike the CONV layer, we note that both (spectral) filtering ratio, and (block-circulant) compression ratio can impact the privacy level and utility, which we study in Section \ref{sec:experiment}.

\vspace{-5pt}
\subsection{Integrate \ourmethod into a DL model}\label{sec:DPtogether}
\vspace{-5pt}
In this section, we show how to apply \ourmethod to train a general deep learning model consisting of many such layers.
Specifically, at each training iteration, \ourmethod computes the per-sample spectral gradient $G^F$, bounds $G^F$ using $L_2$ norm clipping, and adds noise to the spectral gradient using Gaussian mechanism. Then \ourmethod applies spectral filtering to each layer. Without loss of generality, we present the detailed procedure of \ourmethod learning in Algorithm~\ref{alg:PutInDL}.

\begin{algorithm}[t]
    \caption{Training algorithm of \ourmethod in a deep learning model}
    \label{alg:PutInDL}
  \begin{algorithmic}[1]
    \REQUIRE A model with $L$ layers, model parameters $W$, clipping bound $\{C_l\}_l^L$, $\sigma$, filtering parameter $\rho\in(0,1)$, training samples $\{x_i,y_i\}_{i=1}^N$, batch size $B$, total training epochs $T_e$, cost function $J$ learning rate $\alpha$ \\
    \textbf{Output:} Model parameters after $T_e*N/B$ training iterations  $\hat{W}_{T_e*N/B}$ \\
    \FOR{$t\in[T_e*N/B]$}
    \STATE Sample a mini-batch of training samples $\{x_b,y_b\}_{b=1}^B$ by selecting each $\{x_i,y_i\}$ independently with probability $\frac{B}{N}$ using SGM.
    \STATE \textbf{\textit{Stage I:}} Noise addition\\
    \FOR {$b\in 1,2,\cdots,B$}
    \STATE \textbf{Compute per-sample spectral gradient:} \\$G^F(x_b,y_b)=\{G^F_{W_t^{<l>}}(x_b,y_b)\}_l^L$.\\
    \FOR {$l\in 1,2,\cdots,L$}
    \STATE \textbf{$L_2$ norm of clipping:} $\bar{G}^F_{W_t^{<l>}}(x_b,y_b)=G^F_{W_t^{<l>}}(x_b,y_b)/\max\{1,\frac{\|G^F_{W_t^{<l>}}(x_b,y_b)\|_2}{C_l}\}$
    \ENDFOR
    \ENDFOR
    \STATE ${G}_{sum}^F=\sum^{B}_{b=1}\bar{G}^F(x_b,y_b)$
    \STATE \textbf{Gaussian mechanism:} \\$\hat{G}_{sum}^F= Gauss({G}_{sum}^F,C,\sigma)$, where $C=\sqrt{\sum_{l=1}^{L}C_l^2}$ \\
    \textbf{\textit{Stage II:}} Pruning and Inverse Fourier Transform\\
    \FOR {$l\in 1,2,\cdots,L$}
    \STATE \textbf{Spectral filtering and Inverse Fourier Transform:} $\tilde{G}_{sum}^F\leftarrow \mathcal{F}^{-1}(filtering(\hat{G}_{sum}^F))$ 
    \ENDFOR
    \STATE \textbf{Gradient descent}
    $\hat{W}_{t+1}\leftarrow W_{t}-\alpha \frac{1}{B}\tilde{G}_{sum}^F$
    \ENDFOR\\
  \end{algorithmic}
\end{algorithm}

At each training iteration $t$, the mini-batch $\{x_b,y_b\}_{b=1}^B$ is sampled using the Sampled Gaussian mechanism (SGM) \cite{2019arXiv190810530M}. In practical implementation, \ourmethod clips each $g_l$ by a different clipping norm $C_l$.
The $L_2$ norm of $G^F$ can be computed by $C=\sqrt{\sum_{l=1}^{L}C_l^2}$. Based on this clipping strategy, the noise scale in Gaussian mechanism is proportional to $C$ instead of $C_l$--the $L_2$ norm of each layer's gradients. This ensures that the perturbed gradients of all parameters have the same privacy level. In our experiments, we set equal $C_l$ and study the impact of the clipping norm in Section \ref{sec:experiment}. 

In Corollary \ref{cor:composition}, we leverage the RDP based privacy accountant as described in Section \ref{subsec:Background_DPSGD} to compute the overall differential privacy across $T$ epochs. 

\begin{restatable}{corollary}{CompositionCor}
\label{cor:composition}
     Algorithm \ref{alg:PutInDL} achieves $((T_e*N/B)\epsilon+\frac{\log(1/\delta)}{\alpha-1},\delta)-$DP if $\sigma=\frac{\sqrt{2\log(1.25/\delta)}}{\epsilon'}$ where $\epsilon'=\epsilon+\frac{\log(1/\delta)}{\alpha-1}$.
     \vspace{-5pt}
\end{restatable}
\begin{proof}
    The detailed proof is provided in Appendix \ref{Appendix:proof_composition}.
\end{proof}

\section{Experiment}\label{sec:experiment}

\subsection{Experiment setup}\label{subsec:exper_setup}

\noindent \textbf{Experimental Environment.} 
We use Pytorch~\cite{paszke2019pytorch} to implement \ourmethod and 
DP-SGD~\cite{dpsgd}.
For a fair comparison, we follow the provided codes of Opacus~\cite{opacus} to build and train the models for DP-SGD.
All experiments are conducted on a Linux PC with AMD Ryzen Threadripper Pro 3975WX 32-Core Processor, 256 GB memory and NVIDIA GeForce RTX 3090 GPU with 24 GB graphic memory.

\noindent \textbf{Datasets.} We evaluate the proposed \ourmethod training on
public image classification datasets MNIST, CIFAR10 and CIFAR100. 
MNIST~\cite{lecun1998gradient}
consists of 70,000 images of 28 × 28 handwritten grayscale digit,
60,000 images for training and 10,000 for testing. CIFAR10~\cite{krizhevsky2009learning} (32 $\times$ 32 RGB) has 50,000 training images and 10,000 testing images from 10 classes. 
CIFAR100~\cite{krizhevsky2009learning} (32 $\times$ 32 RGB) contains 100 classes, each with 500 training images and 100 test images.

\noindent \textbf{Models.}
We evaluate the model using two training schemes.
In the first scheme, training from scratch, we apply block circulant matrix and perform DP training on all layers. We choose four neural networks with different structures. The first model (\texttt{Model1}) follows the structure evaluated in~\cite{blockcirculant_1}, which contains 4 FC layers with 2048, 1024, 160 and 10 neurons.
The second model (\texttt{Model2}) uses a similar architecture as~\cite{papernot2021tempered}, consisting of 5 CONV layers with Tanh ($\cdot$) activation function. The detailed structure is shown in Table~\ref{tab:conv-1} in Appendix \ref{appendix:model_architecture}.
We further use the LeNet-5~\cite{lecun1998gradient} for MNIST and a model (\texttt{Model3}) that is a similar variant of \texttt{Model2} with 6 convolutional layers and 2 FC layers for CIFAR10 as details shown in Table~\ref{tab:conv-2} in Appendix \ref{appendix:model_architecture}. 
In the second scheme, transfer learning, we load a ResNet-18~\cite{he2016deep} model that is trained over a public dataset (ImageNet~\cite{krizhevsky2012imagenet}) and perform transfer training on CIFAR10.
Furthermore, we use a ResNeXt-29\cite{xie2017aggregated} for transfer learning from CIFAR100 to CIFAR10, following the settings in \cite{tramer2021differentially}. We also follow the state-of-the-art work~\cite{de2022unlocking} by using a Wide-ResNet (WRN-28-10) model~\cite{zagoruyko2016wide} pretrained on down-sampled 32$\times$32 ImageNet images~\cite{chrabaszcz2017downsampled} to perform the transfer learning on CIFAR10 and CIFAR100 datasets.
DP training is applied to the predefined trainable layers.

\noindent {\bf Evaluation metrics.}
\textit{Testing accuracy:} 
the accuracy of a DP trained method on the testing set.
A good defense should obtain the testing accuracy close to that of a model trained without differential privacy.
\textit{Privacy Budget ($\epsilon, \delta$):} 
to measure the privacy constraint of DP training.
We set $\delta=10^{-5}$ for all training and conduct training in two ways. The first way gives a target privacy budget ($\epsilon, \delta$) and sets the training epochs. We report the model accuracy after training. The second way sets the noise scale ($\sigma$) and trains the model for at most 200 epochs, we report the best accuracy epoch and the corresponding accumulated privacy budget ($\epsilon, \delta$).

\noindent \textbf{Parameters setting and general guidelines.}
The key parameter for \ourmethod is the filtering ratio $\rho$ that controls the balance between the DP noise and reconstruction noise. We will discuss the impact of choosing different filtering ratios in Section~\ref{subsec:effectiveness_spectral_dp} and ~\ref{subsec:block_spectral_dp}.  In general, we find that while filtering more coefficients leads to a reduction in the added DP noise, a high filtering ratio results in a greater loss of utility. As a general rule, we  keep at least 50\% of the coefficients in the frequency domain. For more complex tasks, a smaller filtering rate (leaving more coefficients to add noise) may lead to a better utility-privacy tradeoff.

Block size is another hyperparameter used in \ourmethodblock to determine the size of the block circulant matrix used in FC layer.
Similar to existing work~\cite{ding2019req, cirCNN_Ding,dong2020exploring}, 
specifically, we set the block size of the final FC layer as 10, and other layers as 8 for all models. 
We set a uniform clipping norm to 0.5 for \texttt{Model1}, while other models as 0.1.
For training from scratch models, we set the learning rate as 0.01 for \texttt{Model1} and \texttt{Model2} models, and 0.001 for LeNet-5 and \texttt{Model3}.
The batch size is 500.
For the ResNet-18 model used in transfer learning, we choose a learning rate of 0.001, a clipping bound $C=0.1$, and a filtering ratio $\rho=0.2$ for CONV layers. The batch size is  256. For the last two FC layers with 160 and 10 neurons, block sizes for BCM are 16 and 10 with the number of preserved coefficients $k=8$.

\begin{figure*}[t]
     \centering
     \begin{subfigure}[b]{0.33\linewidth}
         \centering
         \includegraphics[width=\textwidth]{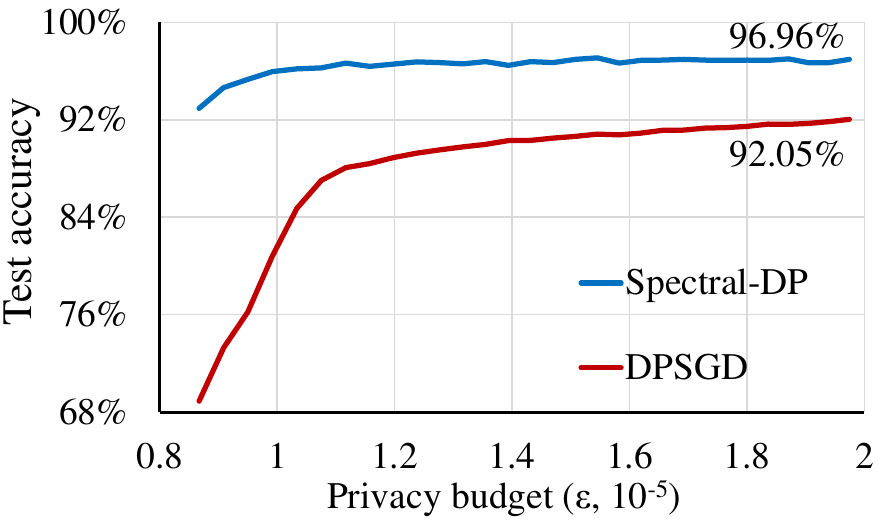}
         \vspace{-15pt}
         \caption{MNIST on \texttt{Model1}}
         \label{subfig:model-1}
     \end{subfigure}
    \begin{subfigure}[b]{0.33\linewidth}
         \centering
         \includegraphics[width=\textwidth]{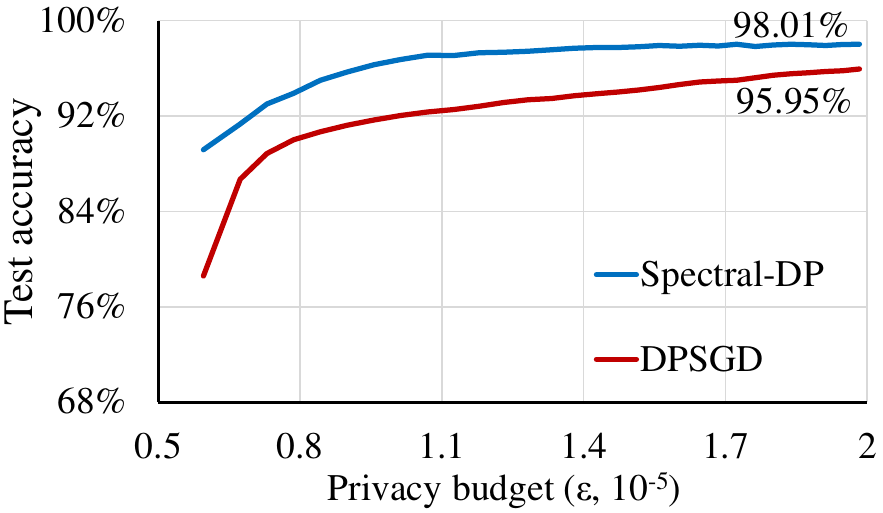}
         \vspace{-15pt}
         \caption{MNIST on LeNet-5}
         \label{subfig:lenet-5}
     \end{subfigure}

     \begin{subfigure}[b]{0.32\linewidth}
         \centering
         \includegraphics[width=\textwidth]{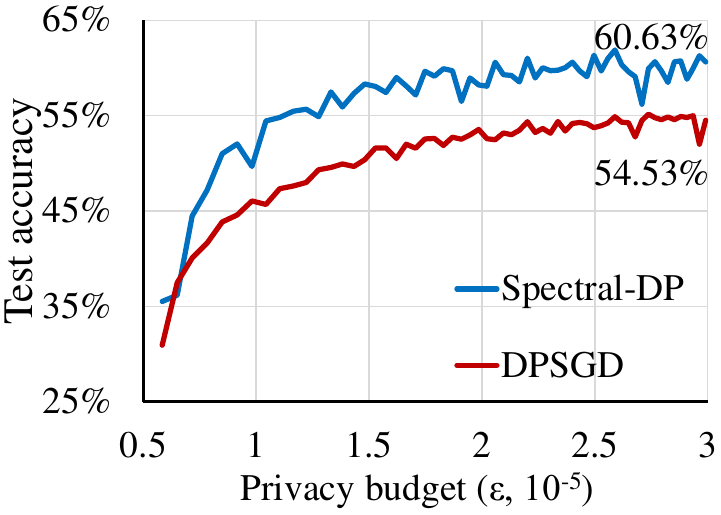}
         \vspace{-15pt}
         \caption{CIFAR10 on \texttt{Model2}}
         \label{subfig:conv1}
     \end{subfigure}
     \hfill
     \begin{subfigure}[b]{0.32\linewidth}
         \centering
         \includegraphics[width=\textwidth]{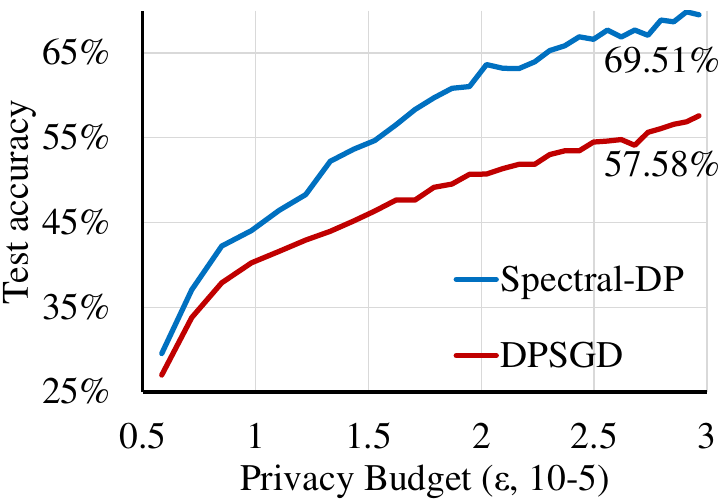}
         \vspace{-15pt}
         \caption{CIFAR10 on \texttt{Model3}}
         \label{subfig:conv2}
     \end{subfigure}
     \hfill
     \begin{subfigure}[b]{0.32\linewidth}
         \centering
         \includegraphics[width=\textwidth]{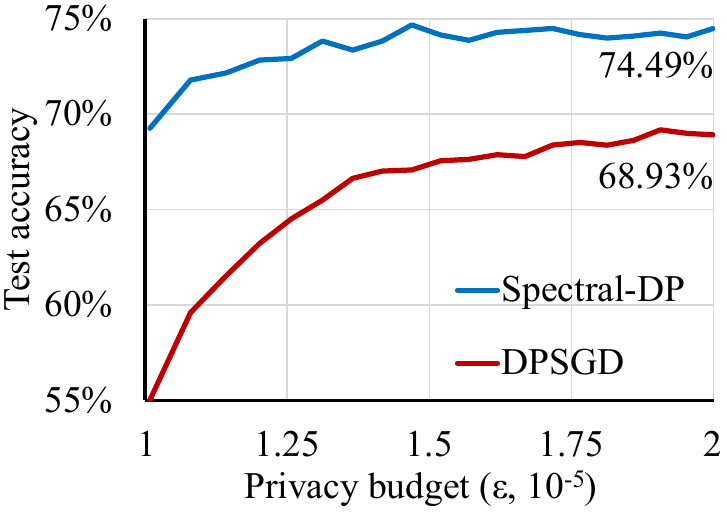}
         \vspace{-15pt}
         \caption{CIFAR10 on ResNet-18 transfer learning}
         \label{subfig:resnet18}
     \end{subfigure}
     \caption{Achieved test accuracy at each privacy budget $(\epsilon, 10^{-5})$ on models and datasets with Spectral-DP and DP-SGD.
     }
     \label{fig:privacy_accuracy}
    \vspace{-10pt}
\end{figure*}

\begin{table}[t]
\centering
\caption{Results for different models with DP-SGD and proposed \ourmethod training.}
\label{tab:best_res}
\resizebox{\linewidth}{!}{%
\begin{tabular}{c|cccc}
\hline
Dataset     & \multicolumn{2}{c|}{MNIST}                                  & \multicolumn{2}{c}{CIFAR10}                    \\ \hline
Model       & \multicolumn{1}{c|}{ \texttt{Model1}} & \multicolumn{1}{c|}{LeNet-5} & \multicolumn{1}{c|}{ \texttt{Model2}}  & \texttt{Model3}          \\ \hline
Privacy budget & \multicolumn{1}{c|}{(2, $10^{-5}$)} & \multicolumn{1}{c|}{(2, $10^{-5}$)} & \multicolumn{1}{c|}{(3, $10^{-5}$)} & (3, $10^{-5}$)  \\ \hline 
Non-Private & \multicolumn{1}{c|}{97.87\%} & \multicolumn{1}{c|}{99.15\%} & \multicolumn{1}{c|}{77.23\%} & 81.22\%           \\ 
DP-SGD       & \multicolumn{1}{c|}{92.05\%} & \multicolumn{1}{c|}{95.95\%} & \multicolumn{1}{c|}{55.15\%} & 57.58\%             \\ 
\ourmethod & \multicolumn{1}{c|}{97.1\%} & \multicolumn{1}{c|}{98.03\%} & \multicolumn{1}{c|}{61.88\%} & 69.51\%           \\ \hline
Max gain & \multicolumn{1}{c|}{34.85\%} & \multicolumn{1}{c|}{13.48\%} & \multicolumn{1}{c|}{19.18\%} &	25.45\% 
     \\ 
Average gain & \multicolumn{1}{c|}{10.70\%} & \multicolumn{1}{c|}{4.49\%} & \multicolumn{1}{c|}{11.91\%} & 19.92\%           \\ \hline
\end{tabular}
}
\vspace{-5pt}
\end{table}

\vspace{-5pt}
\subsection{\ourmethod for Training from Scratch Models}
\vspace{-5pt}
\label{sccratch}
We set target privacy budget to $(2, 10^{-5})$ for MNIST and $(3, 10^{-5})$ for CIFAR10 and train the model for 30 epochs from scratch. 
The test accuracy and privacy budget plots for all models are shown in Figure~\ref{fig:privacy_accuracy}. The best accuracy is reported in Table~\ref{tab:best_res}.
\ourmethod performs well for MNIST tasks on small models and gains more benefits from more sophisticated model training for CIFAR10 tasks. 
Table~\ref{tab:best_res} shows that \ourmethod can maintain accuracy with $\sim1\%$ accuracy drop compared to non-private models on MNIST dataset under privacy budget $\epsilon=2$. 
For CIFAR10 task, we can achieve the accuracy as high as 69.51\% for privacy budget $(3, 10^{-5})$. As a comparison, the state-of-the-art DP training accuracy in~\cite{papernot2021tempered} only delivers 66.2\% accuracy even at a much higher privacy budget--$(7.53, 10^{-5})$. 
We match this accuracy with $\epsilon=2.43$, which is an improvement in the DP-guarantee of $e^{5.1}\approx 164$.

According to Figure~\ref{fig:privacy_accuracy}, we observe that both \ourmethod and DP-SGD exhibit a similar trend, i.e., relaxing privacy constraint increases the accuracy. But \ourmethod always outperforms DP-SGD in all cases.
In most cases, \ourmethod achieves much more accuracy improvement compared to DP-SGD under strict privacy budget constraints. 
In particular, in Figure ~\ref{subfig:model-1}, when $\epsilon=1$, our approach has resulted in 18.75\% accuracy improvements. Overall, \ourmethod leads to $13.48\% \sim 34.85\%$ max accuracy gain and on average $4.49\% \sim 19.92\%$ accuracy gain among all the privacy budget cases compared to DP-SGD. 


\vspace{-5pt}
\subsubsection{Filtering ratio choice of \ourmethod for convolutional layer}
\label{subsec:effectiveness_spectral_dp}


Without loss of generality, we use \texttt{Model2} as an example to evaluate the effectiveness of \ourmethod with different filtering ratio $\rho$. 
We extend the tight privacy budget $(3, 10^{-5})$ evaluated in the previous section with larger $\epsilon$ values (5 and 7) and different filtering ratios on convolutional layers to explore utility improvements in more settings as shown in Table~\ref{tab:conv1-pruning}.
It is intuitive that increasing the filtering ratio decreases the dimension of the additive differentially private noise, but causes more information loss of the weights. We note that there is a tradeoff between the noisy error and the reconstruction error (information loss) which is controlled by the filtering ratio. 
As Table~\ref{tab:conv1-pruning} show, there is a significant accuracy loss for large filtering ratios ($\rho=0.875$) in all cases. 
The models achieve the highest accuracy at $\rho=0.5$ with $\epsilon=5$ and $7$.
With a tight privacy budget of $\epsilon=3$, adding noise causes more prominent utility loss, and filtering more coefficients with $\rho=0.75$ provides better accuracy. 
\textbf{In general, $\rho=0.5$ can be a good starting point to achieve the best utility for convolutional layers.}

\begin{table}[t]
\caption{Test accuracy on CIFAR10 dataset with different filtering ratios for \ourmethod training on \texttt{Model2} models under different privacy budgets.}
\label{tab:conv1-pruning}
\centering
\begin{tabular}{c|ccccc}
\hline
\multirow{2}{*}{\begin{tabular}[c]{@{}c@{}}Privacy\\ Budget\end{tabular}} & \multicolumn{4}{c}{Filtering ratio ($\rho$)} \\ \cline{2-6}
                            & 0       & 0.25    & 0.5              & 0.75    & 0.875   \\ \hline
$(3, 10^{-5})$ & 58.34\% & 58.38\% & 60.26\%          & \textbf{60.63\%} & 55.56\% \\
$(5, 10^{-5})$ & 63.74\% & 63.18\% & \textbf{65.45\%} & 62.20\%  & 55.79\% \\
$(7, 10^{-5})$ & 65.65\% & 65.58\% & \textbf{66.56\%} & 62.28\%   & 58.93\% \\ 
\hline
\end{tabular}
\end{table}



\vspace{-10pt}
\subsubsection{Filtering ratio and block size choice of \ourmethodblock for FC layer}
\label{subsec:block_spectral_dp}

\begin{table}[t]
\centering
\caption{Test accuracy of DP-SGD and \ourmethodblock on MNIST dataset with privacy budget $(2, 10^{-5})$.}
\label{tab:model-1-result}
\begin{tabular}{c|c|cc}
\hline
\multirow{2}{*}{Methods} & \multirow{2}{*}{ \texttt{Model1}} & \multicolumn{2}{c}{Circulant \texttt{Model1}} \\ \cline{3-4} 
                  &                          & BS =8     & BS =16    \\ \hline
Non-Private       & 98.48\%                  & 97.87\%           & 97.38\%           \\
DP-SGD             & 93.55\%                  & 94.77\%           & 95.54\%           \\
\ourmethod        & N/A          & 96.96\%           & 96.85\%           \\ \hline
\end{tabular}
\end{table}

We apply DP-SGD and \ourmethod on \texttt{Model1} under a fixed privacy budget $(2, 10^{-5})$ with various hyper-parameters such as batch size, learning rate, and clipping bound. We report the best test accuracy over all running cases in Table \ref{tab:model-1-result}. 
We apply block circulant matrices to  DP-SGD using the same block sizes as \ourmethod.
We observe an accuracy improvement for DP-SGD trained circulant \texttt{Model1}. 
By using \ourmethodblock for training, we further take advantage of spectral domain based noise reduction and spectral filtering and achieve much better utility than DP-SGD.




We further explore the impact of block size ($BS$) and filtering ratio ($\rho$) using \texttt{Model1}. 
Two different block sizes (8 and 16) under four target differential privacy budgets are evaluated. 
The results under different filtering ratios are shown in Table~\ref{tab:model-1-result-pf}.
Overall, models with $BS = 8$ achieve better utility. The average model accuracy at four target $\epsilon$ across all five filtering ratios, is 94.02\%, 95.78\%, 96.33\%, and	96.57\%, respectively, which is higher than that with $BS = 16$ (92.90\%, 95.29\%, 96.03\%, 96.17\%). This trend is consistent with that of non-private \texttt{Model1}, which is 97.89\% with $BS = 8$ and 97.38\% with $BS = 16$ (in Table~\ref{tab:model-1-result}). 
It indicates that the FC layer often has redundancy and the block circulant matrix can help us further reduce the model size,  
and compressing model weights benefit the spectral calculation of \ourmethodblock. \textbf{Generally, we can adopt a large $BS$ (a power 2 number such as 16) for complex models often containing more redundancy in FC layers, and a small $BS$ (i.e. 8) for small models.}

For filtering ratio ($\rho$), as shown in Table~\ref{tab:model-1-result-pf}, a larger $\rho$ on FC layers leads to better utility in most cases. 
By filtering more frequency coefficients, adding DP noise becomes more smoothly, and the loss due to the large filtering ratio can be compensated. As a result, FC layer with a larger $\rho$ often benefits the utility in \ourmethodblock training. Therefore, \textbf{FC layers in general can have a larger filter ratio than that of convolutional layers ($\rho=0.75$ vs. $\rho=0.5$).}

\begin{table}[t]
\centering
\caption{Test accuracy on MNIST dataset with different filtering ratios and block sizes for \ourmethodblock training on \texttt{Model1} under different privacy budgets.}
\label{tab:model-1-result-pf}
\resizebox{\linewidth}{!}{
\begin{tabular}{c|c|ccccc}
\hline
\multirow{2}{*}{\begin{tabular}[c]{@{}c@{}}Block\\ size\end{tabular}} &
  \multirow{2}{*}{\begin{tabular}[c]{@{}c@{}}Privacy\\ Budget\end{tabular}} &
  \multicolumn{5}{c}{Filtering ratio ($\rho$)} \\ \cline{3-7} 
                    &     & 0       & 0.25    & 0.5              & 0.75 & 0.875 \\ \hline
\multirow{4}{*}{8}  & $(0.5, 10^{-5})$ & 93.08\% & 93.69\% & 93.64\% & \textbf{94.98\%} & 94.71\%\\
                    & $(1.0, 10^{-5})$   & 95.07\% & 95.77\% & 95.57\% & 96.19\% & \textbf{96.29\%} \\
                    & $(1.5, 10^{-5})$ & 95.93\% & 96.24\% & 96.35\% & \textbf{96.60\%}  & 96.52\% \\
                    & $(2.0, 10^{-5})$   & 95.83\% & 96.58\% & 96.59\% & \textbf{96.96\%}  & 96.89\%\\ \hline
\multirow{4}{*}{16} & $(0.5, 10^{-5})$ &  91.57\% & 92.92\% & 93.32\% & 93.30\%  & \textbf{93.41\%} \\
                    & $(1.0, 10^{-5})$   & 95.04\% & 94.08\% & 95.73\% & {95.78\%}  & \textbf{95.80\% }\\
                    & $(1.5, 10^{-5})$ & 95.55\% & 95.76\% & \textbf{96.46\%} & 96.21\% & 96.15\% \\
                    & $(2.0, 10^{-5})$   & 96.02\% & 95.81\% & 96.41\% & \textbf{96.85\%} & 95.74\% \\ \hline
\end{tabular}
}
\end{table}

\vspace{-5pt}
\subsection{\ourmethod in Transfer Learning}
\vspace{-5pt}
\label{subsec:transfer}
\subsubsection{ResNet-18}
In this section, we evaluate the performance of the proposed \ourmethod training on the transfer learning setting. Specifically, we select three transfer training models with different numbers of trainable layers from the bottom of the pretrained ResNet-18. We set the noise scale $\sigma=0.9$ and train the models for 100 epochs, the best accuracy epoch and corresponding privacy budget are reported in Table~\ref{tab:transfer_multi_layer}. 
The 1FC layer model (\texttt{Transfer1}) follows previous work ~\cite{yu2019differentially, dpsgd}--retraining only a hidden layer with 1000 units and a softmax layer with differential privacy.
The \texttt{Transfer2} consists of the last 2 CONV layers and 2 FC layers of the model to be trained. \texttt{Transfer3} further increases the trainable layers to the last 4 CONV layers and 2 FC layers. The baseline accuracy of non-private model increases from 62.31\% to 75.94\% as we increase the number of trainable layers on transfer learning models. 

\ourmethod can benefit more from the increasing number of trainable layers. When increasing to 6 trainable layers in \texttt{Transfer3}, \ourmethod achieves model accuracy close to the non-private model (75.32\% vs 75.94\%) at a small privacy budget ($\epsilon$=2.15). In contrast, the gain of DP-SGD is limited. It suffers more accuracy degradation ($6.61\%$) even with lower privacy guarantee ($\epsilon$=2.89). This clearly indicates that our \ourmethod works much better than DP-SGD when protecting more layers' weights for better privacy is needed. 
\textbf{This further highlights the key advantage of our \ourmethod--better preserving model utility than DP-SGD especially for training models from the scratch with a high-level privacy requirement}, as validated in Section~\ref{sccratch}. Since we obtain the best accuracy in \texttt{Transfer3}, we conduct the following experiments on this model.


\begin{table}[t]
\caption{Transfer learning results for non-private model, DP-SGD training and \ourmethod training with different number of trainable layers. }
\label{tab:transfer_multi_layer}
\centering
\resizebox{\linewidth}{!}{
\begin{tabular}{c|cc|cc|cc}
\hline
Model & \multicolumn{2}{c|}{\texttt{Transfer1}}               & \multicolumn{2}{c|}{\texttt{Transfer2}}        & \multicolumn{2}{c}{\texttt{Transfer3} }       \\ \hline
               & \multicolumn{1}{c|}{$\epsilon$} & Test acc & \multicolumn{1}{c|}{$\epsilon$} & Test acc & \multicolumn{1}{c|}{$\epsilon$} & Test acc \\ \hline
Non-private    & \multicolumn{1}{c|}{$\infty$}       & 62.31\%       & \multicolumn{1}{c|}{$\infty$}       & 70.08\%       & \multicolumn{1}{c|}{$\infty$}       & 75.94\%       \\ 
DP-SGD          & \multicolumn{1}{c|}{3.88}    & 60.10\%       & \multicolumn{1}{c|}{3.24}    & 66.47\%       & \multicolumn{1}{c|}{2.89}    & 69.33\%       \\ 
\textbf{\ourmethod} &
  \multicolumn{1}{c|}{\textbf{2.11}} &
  \textbf{59.11\%} &
  \multicolumn{1}{c|}{\textbf{2.11}} &
  \textbf{70.75\%} &
  \multicolumn{1}{c|}{\textbf{2.15}} &
  \textbf{75.32\%} \\ \hline
\end{tabular}
}
\vspace{-5pt}
\end{table}

\begin{figure}[t]
    \centering
    \includegraphics[width=0.9\linewidth]{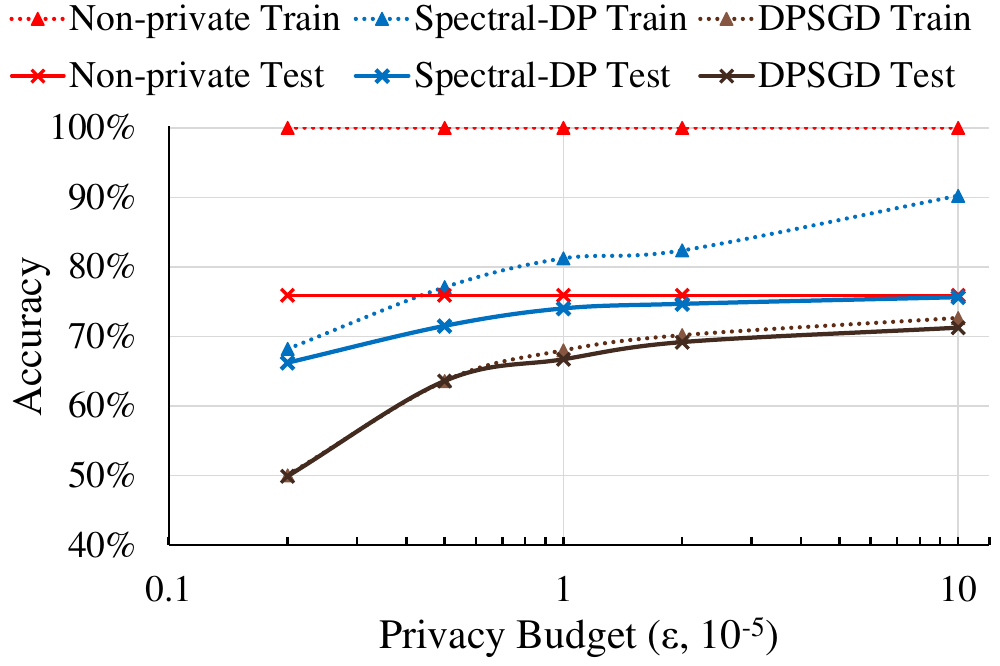}
    \vspace{-5pt}
    \caption{Model accuracy under different privacy budget ($\epsilon, 10^{-5}$) for CIFAR10 transfer models.}
    \label{fig:resnet18}
\end{figure}

\ourmethod can always provide a better tradeoff between utility and privacy than DP-SGD in differential private transfer learning. Figure~\ref{fig:resnet18} shows the results of the \texttt{Transfer3} transfer training with different target privacy budgets. \ourmethod always provides higher accuracy than DP-SGD, from the case with strict privacy constraint (small $\epsilon$) to cases with relaxed privacy requirements. Under an extreme privacy constrain (e.g. $\epsilon$=0.5), our method only causes a 4.41\% accuracy drop compared to the non-privacy model (71.52\% vs 75.94\%), while DP-SGD only achieves 63.60\%, yielding a 12.33\% accuracy loss.

\subsubsection{ResNeXt-29}
\label{subsec:transfer_resnext}
In addition to the layer-wise exploration on ResNet-18, we also adapt our \ourmethod  for comparison according to the state-of-the-art transfer learning setup.
We consider the setting proposed in \cite{tramer2021differentially} for transfer learning from CIFAR100 to CIFAR10. An FC layer-based model is trained on features that are extracted from a ResNeXt\cite{xie2017aggregated} model trained on CIFAR100. DP-SGD and \ourmethod are implemented on the FC layer-based model. Our results are reported in Table \ref{tab:apdx-transfer}. We also compare our results with a DP-SGD utility improvement method \cite{luo2021scalable} that uses transfer learning. Overall, \ourmethod outperforms the DP-SGD trained models in both \cite{tramer2021differentially} and \cite{luo2021scalable} across different privacy budgets $\epsilon$.


\begin{table}[t]
\caption{ResNeXt-29 for transfer learning on CIFAR10}
\label{tab:apdx-transfer}
\resizebox{\linewidth}{!}{
\begin{tabular}{c|ccccc}
\hline
$\epsilon$                                                    & 0.5     & 1.0     & 1.5     & 2.0    &  $\infty$ \\ \hline
DP-SGD in \cite{tramer2021differentially} & -       & -       & -       & 80.00\% & 84.00\%               \\
\ourmethod                                  & \textbf{80.29\%}  & \textbf{80.81\%}  & \textbf{81.71\%} & -       & 84.00\%               \\
\cite{luo2021scalable}                    & 73.28\% & 76.64\% & 81.57\% & -       & 94.10\%               \\ \hline
\end{tabular}
}
\vspace{-5pt}
\end{table}

\subsubsection{WRN-28-10}
\label{subsec:WRN_transfer}
We further evaluate \ourmethod using a pretrained WRN-28-10 model from the down-sampled ImageNet32 dataset to perform transfer training on CIFAR10 and the more complex CIFAR100 datasets. As presented in Table~\ref{tab:cifar10_WRN} and Table~\ref{tab:CIFAR100_WRN}, we retrain the classifier of the WRN model for 20 epochs with different privacy budgets and compare the results with several state-of-the-art works~\cite{de2022unlocking,yu2021do,tramer2021differentially}. The 1 FC layer setting retrains the last layer with 1000 units, while the 2 FC layers setup retrains the whole classifier layer.
Our results show a similar trend--the utility of DP training improves as the number of training layers increases. \ourmethod achieve 94.85\% and 77.52\% at $\epsilon=1$ in 2 FC layers setting for CIFAR10 and CIFAR100, which is higher than the retraining results (93.36\% and 75.99\%) under a relaxed budget $\epsilon=4$ in 1 FC layer training setting.

Our \ourmethod achieves higher accuracy with a strict privacy budget ($94.85\%$ with $\epsilon=1$) compared to other works with relaxed budgets ($94\%$ with $\epsilon=4$~\cite{de2022unlocking}, $94.80\%$ with $\epsilon=2$~\cite{yu2021do} and $92.70\%$ with $\epsilon=2$~\cite{tramer2021differentially}). Even when compared to the strongest setting in \cite{de2022unlocking} (fine-tuning all layers), \ourmethod can achieve similar accuracy with $\epsilon=1$ on CIFAR10 (94.8\%) while exhibiting much higher accuracy (77.52\% with $\epsilon=1$)  on the more complex 100-class dataset–CIFAR100 than that of \cite{de2022unlocking} (e.g. 74.7\% at a relaxed privacy budget $\epsilon=2$). 
\textbf{These results demonstrate that \ourmethod can achieve better utility on more complex datasets and large models as well. }

\begin{table}[t]
\caption{WRN-28-10 transfer learning on CIFAR10}
\label{tab:cifar10_WRN}
\centering
\begin{tabular}{c|lll}
\hline
$\epsilon$                   & \multicolumn{1}{c}{1}       & \multicolumn{1}{c}{2}       & \multicolumn{1}{c}{4} \\ \hline
\ourmethod (2 FC layers) & \textbf{94.85\%}           & \textbf{95.11\%  }                   & \textbf{95.33\% }              \\
\ourmethod (1 FC layer)  & 93.19\%                     & 93.24\%                     & 93.36\%               \\ \hline
DeepMind (2022)~\cite{de2022unlocking}                & 93.10\%                     & 93.60\%                     & 94.00\%               \\
GEP (2021)~\cite{yu2021do}        & 94.30\%          & 94.80\% & -       \\
Feature extraction (2021)~\cite{tramer2021differentially}   & -                & 92.70\% & -       \\ \hline
\end{tabular}
\end{table}

\begin{table}[t]
\caption{WRN-28-10 transfer learning on CIFAR100}
\label{tab:CIFAR100_WRN}
\centering
\begin{tabular}{c|c|ccc}
\hline
                             & $\epsilon$          & 1       & 2       & 4       \\ \hline
\multirow{2}{*}{Spectral-DP} & 2 FC layers       & \textbf{77.52\%} & \textbf{77.78\%} & 78.03\% \\
                             & 1 FC layer          & 74.42\% & 75.65\% & 75.99\% \\ \hline
\multirow{2}{*}{DeepMind~\cite{de2022unlocking} }    & Classifier layer & 70.30\% & 73.90\% & 76.10\% \\
                             & All layers       & 67.40\% & 74.70\% & \textbf{79.20\%} \\ \hline
\end{tabular}
\vspace{-5pt}
\end{table}

\vspace{-2pt}
\subsection{ Replace DP-SGD with \ourmethod in DP-SGD based Existing Works}
\begin{figure}[t]
    \centering
    \includegraphics[width=0.85\linewidth]{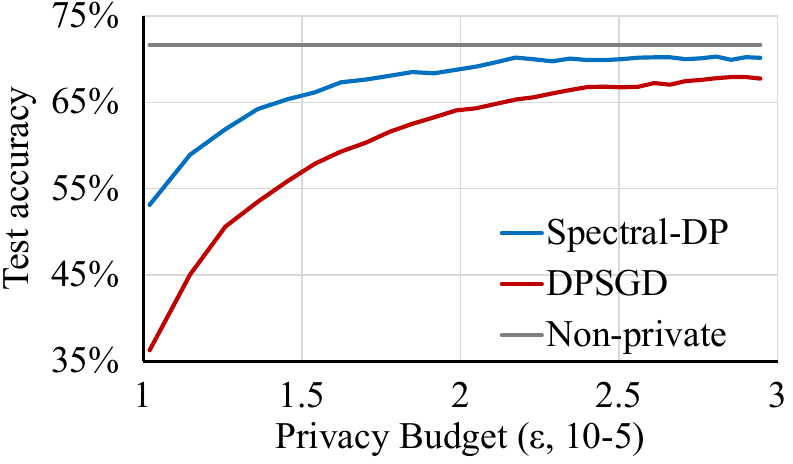}
    \vspace{-5pt}
    \caption{Test accuracy and target privacy budget ($\epsilon, 10^{-5}$) for ScatterCNN model on CIFAR10.}
    \label{fig:scatternet_cifar}
\vspace{-5pt}
\end{figure}

\vspace{-2pt}
Since \ourmethod is proposed as an alternative algorithm to DP-SGD, techniques orthogonal to DP-SGD can be integrated into it as well. Therefore, we can easily replace DP-SGD with \ourmethod in the DP-SGD based existing frameworks, to achieve further utility improvement.
Specifically, we combine a state-of-the-art work~\cite{tramer2021differentially} with our \ourmethod in the training to show the scalability of our approach. 
We adopt the same setting from \cite{tramer2021differentially} that uses the default Scattering Network (ScatterNet) of depth two with wavelets rotated along eight angles from \cite{oyallon2015deep} as a feature extractor to preprocess each data sample. We also apply the same data normalization from ~\cite{tramer2021differentially} on top of the ScatterNet features to obtain the best utility. 
With a target differential privacy budget of ($3, 10^{-5}$), we conduct a grid-search on hyperparameters and report the best results in Table~\ref{tab:best_res_scatter}.
Here ScatterLinear and ScatterCNN adopt similar architectures used in \cite{tramer2021differentially}.
We can observe that on MNIST, \ourmethodblock outperforms DP-SGD on ScatterLinear with accuracy close to that of the non-private model. For CIFAR10, we show the training results on ScatterCNN, and CNN represents the DP training results on \texttt{Model3} as a baseline. We find that with \ourmethod training, we obtain higher accuracy than that of DP-SGD on ScatterCNN. In addition, training with \ourmethod on ScatterNet improves the accuracy by 1.42\% compared to our CNN result, and the accuracy gap is less than 1\% compared with the non-private ScatterCNN result. We also show the test accuracy and privacy budget plot for DP-SGD and \ourmethod in Figure~\ref{fig:scatternet_cifar}. The results indicate that \ourmethod can always outperform DP-SGD and has more gains under tighter privacy budgets.

\begin{table}[]
\caption{Testing accuracy on ScatterNet based model with DP-SGD and proposed \ourmethod training with privacy budget ($3, 10^{-5}$).}
\label{tab:best_res_scatter}
\centering
\begin{tabular}{c|c|cc}
\hline
\multirow{2}{*}{Methods} & MNIST         & \multicolumn{2}{c}{CIFAR10} \\ \cline{2-4} 
                         & ScatterLinear & ScatterCNN     & CNN        \\ \hline
Non-private              & 99.10\%       & 71.68\%        & 81.22\%    \\
DP-SGD                    & 97.66\%       & 67.77\%        & 57.58\%    \\
\ourmethod             & 98.63\%       & 70.93\%        & 69.51\%    \\ \hline
\end{tabular}
\vspace{-5pt}
\end{table}

\vspace{-3pt}
\subsection{Ablation Study}\label{subsec:ablation}
\vspace{-3pt}

\subsubsection{Training from scratch setting}\label{subsec:ablation}

We further analyze how different choices of clipping norm, batch size, and learning rate, impact model performance based on \texttt{Model3} and CIFAR10 dataset. The DP budget is set to be $(\epsilon=3.0,\delta=10^{-5})$, and the model is trained for 30 epochs.
     
     
    

\begin{figure}[t]
    \centering
     \includegraphics[width=0.85\linewidth]{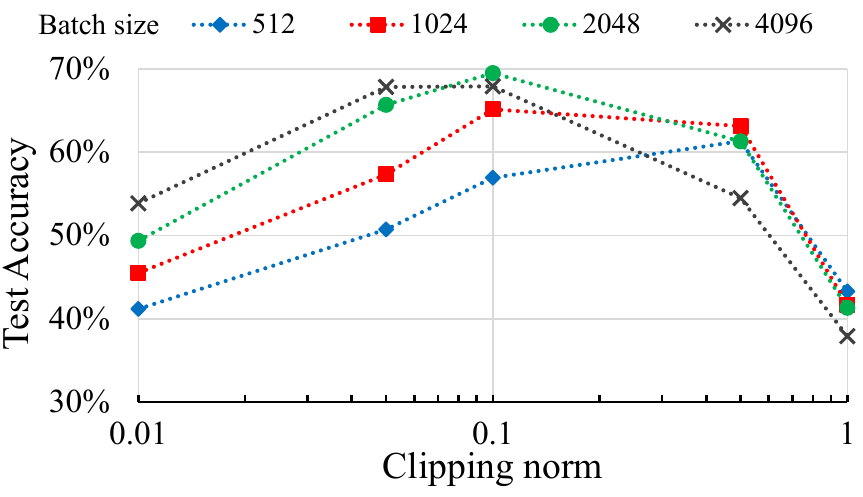}
     \vspace{-5pt}
     \caption{Impact of batch size ($B$) and clipping norm on test accuracy on CIFAR10 dataset.}
     \label{subfig:ablation_BS_Norm}
\vspace{-5pt}
\end{figure}


\textbf{Impact of clipping norm.} Figure \ref{subfig:ablation_BS_Norm} shows how clipping norm ($C$) impacts the differentially private learning using \ourmethod. We observe that the large clipping norm degrades the model utility.
Since the scale of DP noise is proportional to the clipping norm, increasing the norm constraint causes increased gradient noise.
However, too small clipping norms can lead to a large utility loss. This is because the gradient may turn out to be in the opposite direction of the true gradient if the clipping norm is set too low. 
This phenomenon is consistent with the description of the original DP-SGD\cite{dpsgd}. To ensure an efficient \ourmethod private learning, \textbf{we recommend choosing an appropriate clipping norm (beginning with $C=0.1$).} 


\textbf{Impact of batch size.} 
We select 4 batch sizes ($B$) and show the accuracy with each $B$ in Figure~\ref{subfig:ablation_BS_Norm}. We observe that, unlike the non-private model training, $B$ has a relatively large impact on the test accuracy. Changing $B$ from 512 to 2048 leads to 8.17\% accuracy improvement. 
This is because a larger $B$ leads to fewer noise addition iterations. However, the noise scale at a single iteration is positively associated with $B$. When $B$ is too large, the noise has a relatively larger effect than the training iterations. 
Therefore, an appropriate $B$ is essential to balance the utility and the noise addition.

We also find that $B$ and the clipping norm jointly affect the accuracy. When $B=512$, the best accuracy is achieved with $C>0.1$. Meanwhile, when $B=4096$, the best accuracy is achieved by choosing $C<0.1$.  
For \ourmethod training with a fixed privacy budget, a larger batch size implies a larger noise size, while a smaller clipping norm can reduce the noise in the gradient. Therefore, a larger $B$ is compacted with a relatively smaller clipping norm.
As the batch size decreases, the noise scale decreases accordingly. In this case, it requires a relatively large clipping norm to contain as much gradient information as possible.
\textbf{We recommend starting \ourmethod private learning with a relatively large batch size ($B=2048$) and a relatively small clipping norm ($C=0.1$). 
}

\begin{figure}[t]
    \centering
     \includegraphics[width=0.85\linewidth]{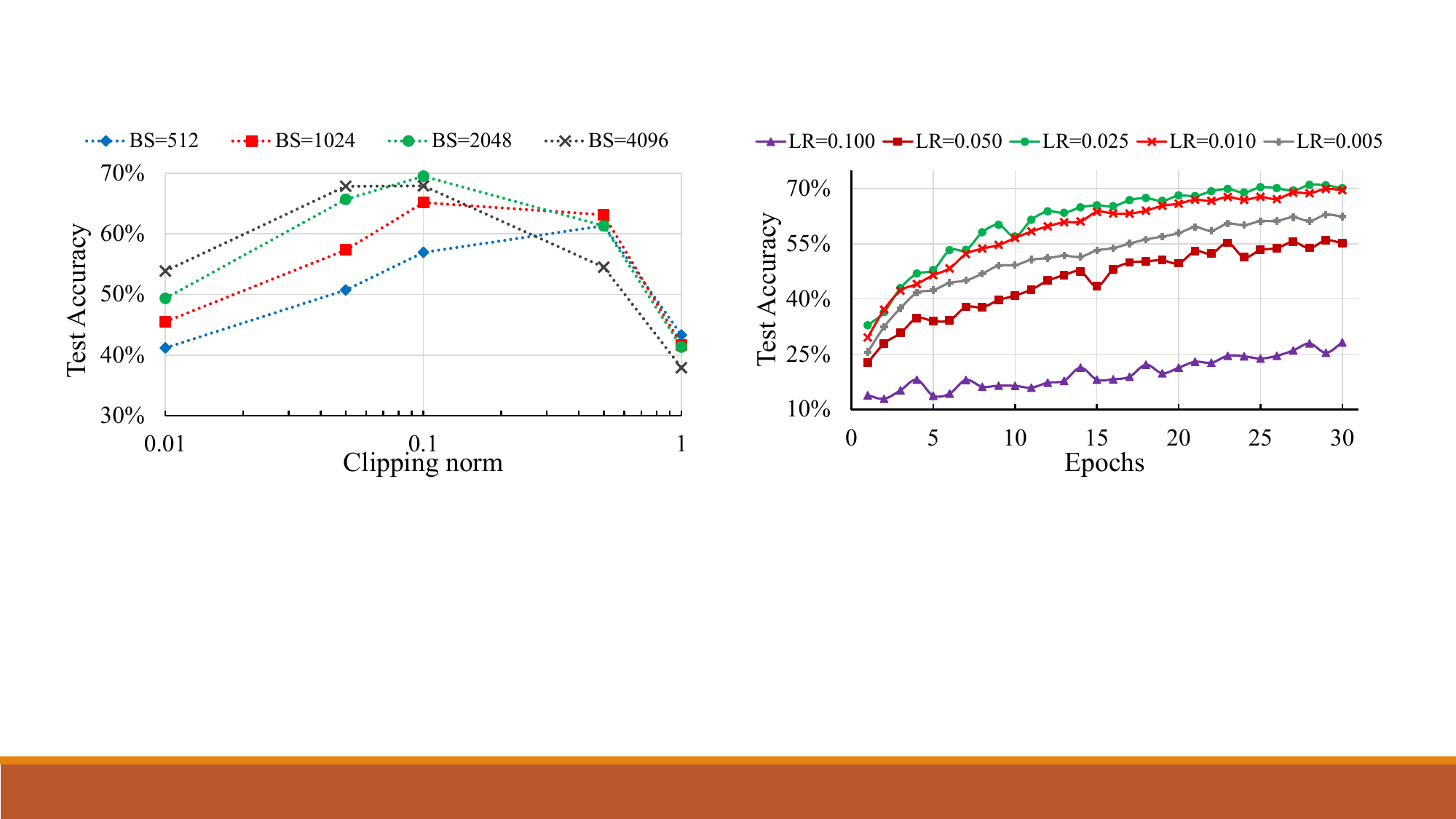}
     \vspace{-5pt}
     \caption{Test accuracy with different learning rate ($LR$) on CIFAR10 dataset.}
     \label{subfig:ablation_LR}
\vspace{-5pt}
\end{figure}

\textbf{Impact of learning rate.} Based on the study of batch size and clipping norm results, we pick the setting of ($B = 2048$, $C=0.1$) and perform \ourmethod training with different learning rates (LR). Figure~\ref{subfig:ablation_LR} shows the plots of the accuracy trends over the training epochs. Too small LR (LR=0.005) can lead to slow convergence of the model and reduce the accuracy over the target training epoch. A large LR=0.1 may also undermine the accuracy significantly. Since a larger LR boosts the weight noise, leading to a random gradient direction that hurts the training convergence. \textbf{The test accuracy is stable when the learning rate is set within a range of $[0.01,0.025]$. Generally, we can set LR=0.01 for the \ourmethod training.}

\begin{figure}[t]
    \centering
    \includegraphics[width=0.85\linewidth]{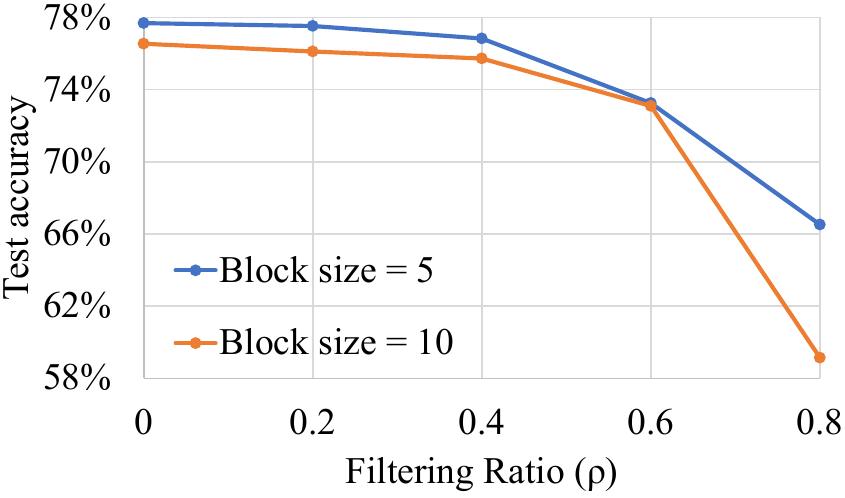}
    \vspace{-5pt}
    \caption{
    Model accuracy with different filtering ratios under privacy budget ($1, 10^{-5}$) with block size 5 and 10 for CIFAR100 transfer models.}
    \label{fig:cifar100_filtering_ratio}
\vspace{-5pt}
\end{figure}

\vspace{-5pt}
\subsubsection{Transfer learning setting}
\label{subsec:transfer_learning_ablation}
In addition to the training from scratch setting, we also discuss the impact of hyperparameters in the transfer learning setting on CIFAR100. In general, the settings of the hyperparameters follow similar trends in the CIFAR10 model for transfer learning, details of which can be found in the Appendix \ref{appdix:ablation_cifar_transfer}.
We retrain the 2 FC layers of the pretrained WRN-28-10 model from ImageNet32 to CIFAR100 by 20 epochs. We adopt different batch sizes, filtering ratios, block sizes, and privacy budgets and summarize the results in Table~\ref{tab:cifar100_ablation}. 

\begin{table*}[h!]
\caption{
Different settings of block size, filtering ratio, privacy budgets,  and batch size on CIFAR100 dataset for WRN-28-10 transfer learning with 2 FC layers.}
\label{tab:cifar100_ablation}
\centering
\resizebox{\linewidth}{!}{
\begin{tabular}{c|ccccccccccccc}
\hline
Block size      & 5  & 5  & 5  & 5  & 5   & 5   & 5   & 5   & 5  & 5  & 5  & 5  & 5  \\
Filtering Ratio ($\rho$)   & 0  & 0  & 0  & 0  & 0.2 & 0.4 & 0.6 & 0.8 & 0  & 0  & 0  & 0  & 0  \\
Privacy Budget ($\epsilon,10^{-5}$)    & 1  & 1  & 1  & 1  & 1   & 1   & 1   & 1   & 2  & 4  & 2  & 4  & 4  \\
Batch size       & 2048    & 1024    & 512     & 256     & 256     & 256 & 256 & 256 & 256     & 256     & 512     & 512     & 2048    \\ \hline
Test Accuracy  & 73.65\% & 75.42\% & 76.59\% & 77.68\% & 77.52\% &   76.83\%  &  73.25\%   &   66.52\%  & 77.78\% & 78.03\% & 76.98\% & 78.00\% & 77.60\%\\ \hline
\end{tabular}
}
\vspace{-10pt}
\end{table*}

The hyperparameter selection for transfer learning does not follow the same trend as that of  training from scratch. First, we compare the impact of different batch sizes from 256 to 2048 with filtering ratio $\rho=0$ and $\epsilon=1$. As Table~\ref{tab:cifar100_ablation} shows, the best accuracy is reached at a batch size of 256, and model accuracy decreases as the batch size increases. Since transfer learning just retrains a small number of parameters, a smaller batch size facilitates more fine-grained learning and ultimately results in improved utility.

Next we explore the impact of filtering ratios. We choose five filtering ratios (0, 0.2, 0.4, 0.6, and 0.8) for the two FC layers with the accuracy shown in Figure~\ref{fig:cifar100_filtering_ratio}. Our findings show that a smaller filtering ratio results in better utility, indicating that
\ourmethodblock can benefit from the model training with noise added in the spectral domain 
and reconstruction error has a more significant impact on model accuracy in transfer learning settings.

Block size is not a dominating factor for model utility in this case. We can observe that from Figure~\ref{fig:cifar100_filtering_ratio}, a smaller block size (5) leads to slightly better model utility, but its influence is not as significant as that of filtering ratios. This also indicates a tradeoff between efficiency and utility as a larger block size means fewer trainable parameters due to higher model compression, but lower model utility.

\vspace{-3pt}
\section{Limitations}\label{sec:limitations}
\vspace{-3pt}
Although \ourmethod effectively reduces the DP noise and achieves better privacy utility tradeoff, 
there is scope for further improvement.
First, while \ourmethod achieves the same level of utility as that of the non-private learning for the transfer learning setup, there still exists the utility or accuracy gap for model training from scratch setup. Second, offering rigorous guidelines to select optimal key parameters such as filtering ratio $\rho$, with theoretical guarantees is still challenging, 
due to the interplays among multiple factors, including dataset characteristics, model complexity, privacy budget, clipping specification, and batch size.
A comprehensive analysis of these factors would be a promising avenue for future research. Third, there are other domain transform methods such as wavelet, etc. that can be explored for further improvement, as Spectral-DP opens up a new era for private deep learning using DP.

\vspace{-3pt}
\section{Related Work}\label{sec:relatedwork}
\vspace{-3pt}

The subject of differential privacy in the context of machine learning attracted significant scientific interest and has been used in support vector machines~\cite{rubinstein2009learning}, linear regression~\cite{chaudhuri2008privacy,zhang2012functional}, and risk minimization~\cite{chaudhuri2011differentially, bassily2014private}.
In recent years, more works have focused on privacy-preserving training for deep learning. Private Aggregation of Teacher Ensembles (PATE)~\cite{papernot2016semi, papernot2018scalable} is one approach that transfers the knowledge from an ensemble of teachers trained on the disjoint subsets of training data to train a student model through the noisy aggregation of teachers' answers.

Differentially private (stochastic) gradient descent (DP-SGD) \cite{dpsgd,yu2019differentially} as described earlier perturbs the gradient at each update with random noise drawn from Gaussian distribution during the training. Some recent works aim to do noise reduction on DP training by adding noise into the reduced gradient.
In \cite{yu2021do}, a gradient embedding perturbation (GEP) is proposed to achieve higher utility by adding noise into a low-dimensional projected gradient.
\cite{yu2021large} designs reparametrized gradient perturbation (RGP), which perturbs the gradients of the low-rank gradient-carrier matrix and reconstructs the update of the original weights from the noisy gradients.
The framework in \cite{nasr2020improving} encodes gradients, mapping them to a smaller vector space, and hence is able to provide DP guarantees for different noise distributions.

These gradient dimension reduction techniques rely on either a projection or decomposition that maps gradients into a smaller subspace. 
We note that the key principle of these methods is gradient approximation, and therefore they would inevitably cause undesired utility loss.
Our work involves performing lossless transformation of gradients using the Fourier transform and applying filtering in the spectral domain to improve privacy utility trade-off.
Fourier Perturbation Algorithm (FPA) is proposed in \cite{FourierDP_database2010} and further optimized in \cite{acs2012differentially} to address the poor performance of conventional differential privacy aggregation algorithm for time-series data. FPA focuses on the differential privacy of time-series data and conducts noise aggregation in the frequency domain, which is similar to \ourmethod.

Several recent works focus on a wealth of areas to improve the utility of the DP-SGD trained models. In Section \ref{sec:introduction}, we discussed tempered sigmoid activations \cite{papernot2021tempered} introduced to help control the gradient norm of the loss function, thus mitigating the negative effects of clipping and noising. 
\cite{luo2021scalable} leverages additional public data transfer learning to minimize the number of trainable parameters in the model to optimize the privacy-utility tradeoff. 
\cite{cheng2022dpnas} proposes a framework to perform a neural architecture search with DP-aware candidate model training to find the suitable model for DP training. 
Work in \cite{tramer2021differentially} as mentioned earlier demonstrates that better features in data can lead to higher utility of DP-SGD trained model.

We note that these works focus on how to improve DP-SGD by modifying the model structure or preprocessing the data. Furthermore, other directions including \ie clipping~\cite{Stevens2022,Andrew2019,Pichapati2019} and privacy budget allocation~\cite{lee2018concentrated,asi2021private, yu2019differentially} do not change the DP-SGD algorithm. In contrast, \ourmethod focuses on the gradient updating algorithm of DP training that adapts spectral domain DP perturbation into deep learning as an alternative to DP-SGD.  





Another emerging topic is the use of differential privacy to protect privacy and robustness in federated learning (FL)~\cite{mcmahan2017communication}. Client-based Differential Privacy has been introduced in~\cite{mcmahan2017learning, geyer2017differentially} in order to hide any information that is specific to a single client’s training data. Noising before model aggregation FL (NbAFL)~\cite{wei2020federated} and LDP-Fed~\cite{truex2020ldp} perturb the trained parameters locally in each client before aggregation to ensure local differential privacy. \cite{stevens2022efficient} proposes a new protocol for differentially private secure aggregation based on techniques from Learning With Errors~\cite{regev2009lattices}. As future work, it would be a good opportunity to adapt and combine our \ourmethod to the FL process to provide better privacy-utility tradeoffs in these complex scenarios.

\vspace{-3pt}
\section{Conclusion}\label{sec:conclusion}
\vspace{-3pt}
In this work, we propose \ourmethod, an alternative to DP-SGD in the context of differentially private deep learning. \ourmethod combines differentially private noise addition in the spectral domain with spectral filtering which enables reduction of  noise scale to improve utility. Our extensive experimental results show that our approach has uniformly better privacy utility tradeoff compared to state-of-the-art methods. Our contribution is a new paradigm to gradient perturbation in the context of deep learning, which can be further built upon. For instance, Fourier is only one example of a unitary transformation, and although it is widely used, other transformations could be considered for spectral perturbation. Likewise, developing alternative weight restructuring, and more general filtering approaches might yield interesting and broader insights into the general principle of spectral domain based methods to achieve differential privacy in deep learning. 

\section*{Acknowledgment}
This research was partially supported by the National Science Foundation through the grants CCF-1617889, CCF-2011236, CCF-2006748, and partially through a Lehigh internal CORE grant.

\bibliographystyle{plain}
\bibliography{refs}

\appendices

\section{}
\subsection{Proof of Theorem \ref{thm:FourierDP}}\label{Appendix:proof_privacythm}
\Privacythm*
\begin{proof}
    The proof relies on the Theorem 3.22 in \cite{DP_algorithm} and the post-processing property of DP algorithm. 
    First, we show that $\tilde{F^N}$ is ($\epsilon,\delta$) if $\sigma=\sqrt{2\log(1.25/\delta)}/\epsilon$. Since the mechanism of obtaining $\tilde{F^N}$ is a Gaussian mechanism, and the variance of the $2S^2\log(1.25/\delta)/\epsilon^2$ where $S$ is the sensitivity of $F^N$, Following the Theorem 3.22 in \cite{DP_algorithm}, $\tilde{F^N}$ is ($\epsilon,\delta$) differentially private under the condition that the $L_2$-sensitive of $F^N$ is $S$. 
    The spectral filtering and the inverse Fourier transformation are the post-processing of $\tilde{F}^N$. Since the post-processing does not change the differential privacy budget, the Fourier DP follows the same privacy budget as the Gaussian mechanism.
\end{proof}
\begin{theorem}{(Theorem 3.22 in \cite{DP_algorithm})}
    Let $\epsilon \in (0,1)$ be arbitrary. For $c^2 > 2ln(1.25/\delta)$, the Gaussian Mechanism with parameter $\sigma \ge c\triangle _2(f)/\epsilon$ is ($\epsilon,\delta$) differentially private.
\end{theorem}

\subsection{Proof of Proposition \ref{prop:noise_reduction}}\label{Appendix:noise_reduction}
\PropNoiseReduction*
\begin{proof}
    Let $V_i^K=P_K(V_i)$, then the resulting $v_n=\text{I-FT}(V_i^K)$. In detail, it can be formulated as
    $$\begin{array}{ccl}
        v_n& =&\frac{1}{\sqrt{N}}\sum_{i=0}^{N-1}V_i^K\cdot e^{\frac{j2\pi}{N}in} \\
        & =&\frac{1}{\sqrt{N}}\sum_{i=0}^{K}V_i\cdot e^{\frac{j2\pi}{N}in}\\
        & =&\frac{1}{\sqrt N}\sum_{i=0}^{K-1}\{V_i\}\cdot\cos(\frac{2\pi}{N}in)\\
        &&+ j\frac{1}{\sqrt N}\sum_{i=0}^{K-1}\{V_i\}\cdot\sin(\frac{2\pi}{N}in)
    \end{array}$$
    Define $c_{n,i}=\frac{1}{\sqrt{N}}\{V_i\}\cdot\cos(\frac{2\pi}{N}in)$, then we have
    $$\frac{1}{\sqrt N}\sum_{i=0}^{K-1}\{V_i\}\cdot\cos(\frac{2\pi}{N}in)=\sum_{i=0}^{K-1}c_{n,i}$$ 
    Following the property of the normal distribution, we have $$\sum_{i=0}^{K-1}c_{n,i}\sim\calN(0,\sum_{i=0}^{K-1}\frac{1}{N}\cdot\cos^2(\frac{2\pi}{N}in)\sigma^2S^2)$$
    Simplifying the variance of the distribution, we have $$\sum_{i=0}^{K}c_{n,i}\sim\calN(\frac{K}{N}\cdot\frac{\sigma^2S^2}{2})$$
    Similarly, we have $\frac{1}{\sqrt N}\sum_{i=0}^{K}\{V_i\}\cdot\sin(\frac{2\pi}{N}in)\sim\calN(\frac{K}{N}\cdot\frac{\sigma^2S^2}{2})$. This indicates that $v_n\sim\mathbb{C}\calN(\frac{K}{N}\cdot\sigma^2S^2)$ has the same scale of $\calN(\frac{K}{N}\cdot\sigma^2S^2)$
\end{proof}
\subsection{Proof of Proposition \ref{prop:noise_reduction}}\label{Appendix:noise_reduction2D}
\PropNoiseReductionConv*
\begin{proof}
    Let $V_{i,j}^K=P^K_{2D}(V_{i,j})$, then the resulting $v_{mn}=\mathcal{F}^{-1}(V_{ij}^K)$. In detail, it can be formulated as
    \begin{small}
    $$\begin{array}{ccl}
        v_{mn}& =&\frac{1}{N}\sum_{i=0}^{K-1}\sum_{j=0}^{K-1}V_{ij}\cdot e^{\frac{j2\pi}{N}im+jn} \\
        & =&\frac{1}{N}\sum_{i=0}^{K-1}\sum_{j=0}^{K-1}\{V_{ij}\}\cdot\cos(\frac{2\pi}{N}(im+jn))+\\
        &&\sqrt{-1}\frac{1}{N}\sum_{i=0}^{K-1}\sum_{j=0}^{K-1}\{V_{ij}\}\cdot\sin(\frac{2\pi}{N}(im+jn))
    \end{array}$$
    \end{small}
    Define $c_{mn,ij}=\frac{1}{N}\{V_{i,j}\}\cdot\cos(\frac{2\pi}{N}(im+jn))$, then we have
    $$\frac{1}{N}\sum_{i=0}^{K-1}\sum_{j=0}^{K-1}\{V_{ij}\}\cdot\cos(\frac{2\pi}{N}(im+jn))=\sum_{i=0}^{K-1}\sum_{j=0}^{K-1}c_{mn,ij}$$ 
    Following the property of the normal distribution, we have 
    \begin{footnotesize}
    $$\sum_{i=0}^{K-1}\sum_{j=0}^{K-1}c_{mn,ij}\sim\calN(0,\sum_{i=0}^{K-1}\sum_{j=0}^{K-1}\frac{1}{N^2}\cdot\cos^2(\frac{2\pi}{N}(im+jn))\sigma^2S^2)$$
    \end{footnotesize}
    Simplifying the variance of the distribution, we have $$\sum_{i=0}^{K-1}\sum_{j=0}^{K-1}c_{mn,ij}\sim\calN(\frac{K^2}{N^2}\cdot\frac{\sigma^2S^2}{2})$$
    Similarly, we have $\sum_{i=0}^{K-1}\sum_{j=0}^{K-1}\{V_{ij}\}\cdot\sin(\frac{2\pi}{N}im+jn)\sim\calN(\frac{K^2}{N^2}\cdot\frac{\sigma^2S^2}{2})$. This indicates that $v_{mn}\sim\mathbb{C}\calN(\frac{K^2}{N^2}\cdot\sigma^2S^2)$ which has the same scale of $\calN(\frac{K^2}{N^2}\cdot\sigma^2S^2)$
\end{proof}
\subsection{Proof of Corollary \ref{cor:composition}}\label{Appendix:proof_composition}
\CompositionCor*
\begin{proof}
    In Algorithm \ref{alg:PutInDL}, let $\sigma=\frac{\sqrt{2\log(1.25/\delta)}}{\epsilon'}$ and $\epsilon'=\epsilon+\frac{\log(1/\delta)}{\alpha-1}$, the Gaussian mechanism guarantees $(\epsilon',\delta)-$DP following Theorem 3.22 in \cite{DP_algorithm}. This is equivalent with $(\alpha,\epsilon)-$RDP according to Proposition \ref{prop:RDP_conversion}.
    
    Proposition \ref{prop:RDP_composition} further shows Algorithm \ref{alg:PutInDL} satisfies $(\alpha,(T_e*N/B)\epsilon)-$RDP. The $(\alpha,(T_e*N/B)\epsilon)-$RDP can be converted back to $((T_e*N/B)\epsilon+\frac{\log(1/\delta)}{\alpha-1},\delta)-$DP based on the Proposition \ref{prop:RDP_conversion}.
\end{proof}

\subsection{Model architectures}\label{appendix:model_architecture}
Here we show the detailed model architectures of the model used in the paper.

\begin{table}[t!]
\centering
\caption{Architecture of model that only consists of Fully Connected (FC) layers. A layer name ending with BC means that the weights matrix of such layer is block circulant.}
\label{tab:model-1}
\begin{tabular}{cc|ccc}
\hline
\multicolumn{2}{c|}{Model-1}  &   \multicolumn{3}{c}{Circulant \texttt{Model1}}      
\\\hline
Layer & Weight & Layer & Weight & Block\\ \hline
FC1            & $784\times2048$  & FC1-BC  & $784\times2048$  & 8/16\\
FC2            & $2048\times1024$ & FC2-BC  & $2048\times1024$ & 8/16\\
FC3            & $1024\times160$  & FC3-BC  & $1024\times160$  & 8/16\\
FC4            & $160\times10$    & FC4-BC  & $160\times10$    & 10\\ \hline
\end{tabular}
\end{table}

\begin{table}[t!]
\centering
\caption{Architecture of \texttt{Model2} model that consists of convolutional layers. 
}
\label{tab:conv-1}

\begin{tabular}{ll}
\hline
\multicolumn{2}{c}{\texttt{Model2}}                           \\ \hline
Layer       & Parameters                             \\ \hline
Convolution & 32 filters of 3x3, stride 1, padding 1 \\
Max-Pooling & 2 × 2, stride 2                        \\
Convolution & 64 filters of 3x3, stride 1, padding 1 \\
Max-Pooling & 2 × 2, stride 2                        \\
Convolution & 64 filters of 3x3, stride 1, padding 1 \\
Max-Pooling & 2 × 2, stride 2                        \\
Convolution & 64 filters of 3x3, stride 1, padding 1 \\
Max-Pooling & 2 × 2, stride 2                        \\
Convolution & 10 filters of 3x3, stride 1, padding 1 \\
Average     & Over spatial dimensions                \\ \hline
\end{tabular}
\end{table}

\begin{table}[t!]
\centering
\caption{The architecture of \texttt{Model3} model for CIFAR10.}
\label{tab:conv-2}
\begin{tabular}{ll}
\hline
\multicolumn{2}{c}{\texttt{Model3}}                                \\ \hline
Layer           & Parameters                              \\ \hline
Convolution x2  & 32 filters of 3x3, stride 1, padding 1  \\
Max-Pooling     & 2 × 2, stride 2                         \\
Convolution x2  & 64 filters of 3x3, stride 1, padding 1  \\
Max-Pooling     & 2 × 2, stride 2                         \\
Convolution x2  & 128 filters of 3x3, stride 1, padding 1 \\
Max-Pooling     & 2 × 2, stride 2                         \\
Fully connected & 120 units                               \\
Fully connected & 10 units                                \\ \hline
\end{tabular}
\end{table}

\begin{table}[t!]
\centering
\caption{The architecture of ScatterCNN model for CIFAR10, with Tanh activations.}
\label{tab:scatterCNN}
\begin{tabular}{ll}
\hline
\multicolumn{2}{c}{ScatterCNN}                                \\ \hline
Layer           & Parameters                              \\ \hline
Convolution   & 64 filters of 3x3, stride 1, padding 1  \\
Max-Pooling     & 2 × 2, stride 2                         \\
Convolution   & 60 filters of 3x3, stride 1, padding 1 \\
Fully connected & 10 units                                \\ \hline
\end{tabular}
\end{table}

\begin{table*}[t]
\caption{Different settings of training epochs, filtering ratios and privacy budgets on CIFAR10 dataset for ResNet-18 transfer learning with 4CONV+2FC trainable layers (\texttt{Transfer3}).}
\centering
\begin{tabular}{c|ccccccccc}
\hline
Training Epochs        & 10      & 10      & 10      & 20      & 40      & 10      & 10      & 10  & 20\\
Filtering Ratio ($\rho$)          & 0.75    & 0.5     & 0.2     & 0.2     & 0.2     & 0.2     & 0.2     & 0.2  & 0.2  \\
Privacy Budget $(\epsilon,10^{-5})$ & 0.5     & 0.5     & 0.5     & 0.5     & 0.5     & 0.2     & 1       & 2 & 2 \\\hline
Train Accuracy & 76.36\% & 77.50\% & 77.68\% & 76.83\% & 74.39\% & 71.44\% & 81.11\% & 83.82\% & 86.55\% \\
Test Accuracy  & 71.50\% & 72.61\% & 72.89\% & 71.98\% & 70.31\% & 68.52\% & 73.72\% & 74.53\% & 74.49\% \\ \hline
\end{tabular}
\label{tab:trans}
\end{table*}

\subsection{Ablation study on CIFAR10 transfer models}\label{appdix:ablation_cifar_transfer}
In this section, we also discuss the impact of hyperparameters in the transfer learning setting based on the \texttt{Model3} and CIFAR-10 dataset. We summarize the results with different pairs of training epochs, filtering ratio ($\rho$) and target privacy budget in Table~\ref{tab:trans}. First, we discuss the impact of training epochs for a target privacy budget training. We set a strict target privacy budget $\epsilon=0.5$ and conduct 10, 20, and 40 epochs of training. More training epochs results in less noise addition to each training step but leads to an increased number of training steps. Different choices of training epochs may affect the convergence and final accuracy of the model. 
As Table~\ref{tab:trans} shows, fewer training epochs leads to higher accuracy when $\epsilon=0.5$. However, when the privacy budget is relaxed to $\epsilon=2$, we observe the similar test accuracy for 10 and 20 epochs.

We thus conduct experiments with 10 training epochs and set $\rho$ to 0.2, 0.5 and 0.75 under target privacy budget ($0.5, 10^{-5}$). Under this tight budget, setting $\rho$ to 0.2 and 0.5 results in the similar model performance. However, unlike the cases in Section~\ref{subsec:effectiveness_spectral_dp} for the training from scratch models, in transfer learning, setting a smaller rate like 0.2 gives better performance. The reason is because during the transfer learning process, we usually only tune the final layers of a pre-trained model, a large filtering ratio indicates losing too much information, thus worse accuracy.

\begin{figure}[h]
    \centering
    \includegraphics[width=0.95\linewidth]{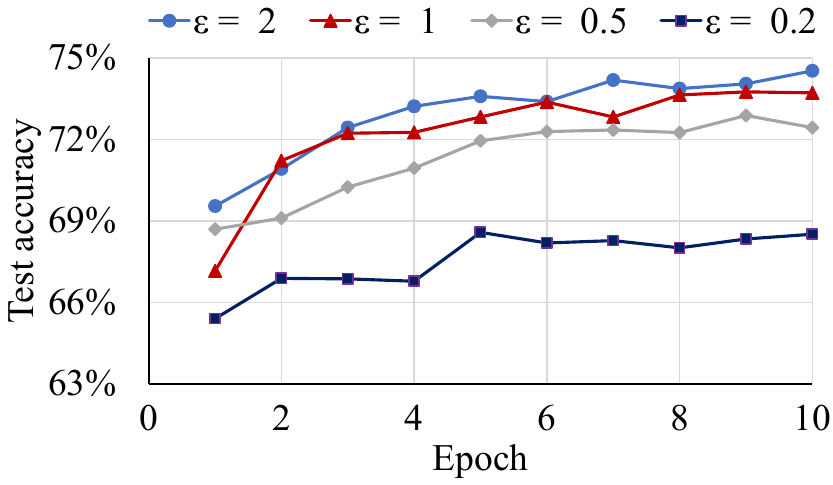}
    \caption{Test accuracy under different target privacy budgets ($\epsilon, 10^{-5}$) for CIFAR10 transfer models.}
    \label{fig:resnet18-dptrain}
\vspace{-10pt}
\end{figure}

Finally, we draw training curves using 10 training epochs with  $\rho=0.2$ under different privacy budgets in Figure~\ref{fig:resnet18-dptrain}. It demonstrates that our \ourmethod allows the model to converge quickly in the first several epochs. When $\epsilon>= 0.5$, the training curves demonstrate the similar tradeoff between privacy budget and accuracy. The utility of the model is limited only when the privacy budget becomes very low ($\epsilon=0.2$). In this case, training after the 5th epoch cannot further improve the accuracy. Nevertheless, this result (68.52\%) is still higher than the DP-SGD trained result (66.72\%) under a more relaxed privacy budget $\epsilon=1$.

\subsection{Discussion on complexity}\label{apdx:complexity}

\subsubsection{Complexity of \ourmethod in CONV layer} We implement the DFT using the Fast Fourier Transform (FFT) algorithm. Given a 2D convolution with signal size $n\times n$ and kernel size $d\times d$, the computational complexity of conventional convolution is $O(n^2d^2)$. For FFT-based convolution, the complexity composes of two parts: First, the complexity of FFT operation is $O(n^2\log n)$ and there are three times for doing the FFT (including FFT of the signal, FFT of the kernel, and inverse FFT of the spectral multiplication). Second, the multiplication operation has a complexity of $4n^2$. Therefore, the complexity of FFT-based convolution is $O(n^2\log n)$. This implies that FFT-based convolutions are more computationally efficient if $log (n)<d^2$.


In Eq. (\ref{eq:conv_grad}), the gradient of $w_{i,j}$ is the convolution of the gradient of $A_I$ and $X_j$. In neural network, most commonly used sizes of $w_{i,j}$ are $3\times 3$ and $5\times 5$. When computing the gradient of such $w_{i,j}$, the size of $X_j$ is close to the size of $A_i$, leading a more efficient convolution using FFT.

\subsubsection{Complexity of \ourmethodblock in FC layer}
An advantage of Block Spectral-DP is that it employs a block circulant matrix to reduce storage complexity. Specifically, for a $d\times d$ block circulant matrix, the storage complexity can be reduced from $O(d^2)$ to $O(d)$ by using a block size of $d$. In addition to this, the use of a block circulant matrix can also result in a reduction in computational complexity. For example, in a fully connected layer with a block circulant matrix-based weight of size $m\times n$, the computational complexity can be reduced from $O(n^2)$ to $O(n\log n)$ \cite{cirCNN_Ding}.

\end{document}